\documentclass[letterpaper]{article}
\usepackage[preprint]{aaai2027}
\usepackage[hyphens]{url}
\usepackage{graphicx}
\usepackage{natbib}
\usepackage{caption}
\usepackage{amsmath}
\usepackage{booktabs}
\newcommand{\vpara}[1]{\vspace{0.05in}\noindent \textbf{#1 }}
\newcommand{\ProjectPageURL}{https://github.com/RUCKBReasoning/RoboRecover}
\newcommand{\GitHubLink}{%
  \leavevmode
  \pdfstartlink attr{/Border [0 0 0]} user{/Subtype /Link /A << /Type /Action /S /URI /URI (\ProjectPageURL) >>}%
  \textcolor{blue}{GitHub}\pdfendlink
}
\title{RoboRecover: Benchmarking Robot Policy Recovery under Execution Deviations}
\author{
    Yang Li\textsuperscript{\rm 1,\rm 2},
    Chen Zhao\textsuperscript{\rm 1,\rm 2},
    Zhuoran Wang\textsuperscript{\rm 1,\rm 2},
    Jiankang Wang\textsuperscript{\rm 3},
    Chao Shao\textsuperscript{\rm 1,\rm 2},
    Yihan Lin\textsuperscript{\rm 1,\rm 2},
    Haitao Shen\textsuperscript{\rm 1,\rm 2},
    Jing Zhang\textsuperscript{\rm 1,\rm 4}\corresponding
}
\affiliations{
    \textsuperscript{\rm 1}School of Information, Renmin University of China, Beijing, China\\
    \textsuperscript{\rm 2}Key Laboratory of Data Engineering and Knowledge Engineering, Beijing, China\\
    \textsuperscript{\rm 3}University of Science and Technology of China\\
    \textsuperscript{\rm 4}Engineering Research Center of Database and Business Intelligence, Beijing, China
}

\begin{document}
\maketitle
% Canonical manuscript values, kept as plain LaTeX for Overleaf editing.
% Change only the second argument of a command. Rates are percentages unless
% a comment says otherwise. Detailed tables and analysis code remain outside
% the paper repository under paper_maintenance/{data,scripts}/roborecover.

% Corpus and fixed splits (RT = RoboTwin, LB = LIBERO).
\newcommand{\TotalRecords}{2{,}000}
\newcommand{\PlatformRecords}{1{,}000}
\newcommand{\TrainRecords}{800}
\newcommand{\TestRecords}{200}
\newcommand{\NumPolicies}{6}
\newcommand{\RTTestTraj}{183}
\newcommand{\RTTestTasks}{34}
\newcommand{\LBTestTasks}{38}
\newcommand{\LBNatural}{42}
\newcommand{\LBPerturbed}{127}
\newcommand{\LBHuman}{31}

% RoboTwin: Seed = matched seed-only; RSR = replay Overall; Lo/Hi =
% trajectory-cluster 95% CI; Macro = equal-weight Macro-9 point estimate.
\newcommand{\RTXSeed}{54.33}
\newcommand{\RTXRSR}{49.83}
\newcommand{\RTXLo}{43.25}
\newcommand{\RTXHi}{56.35}
\newcommand{\RTXMacro}{40.65}

\newcommand{\RTLingVLASeed}{64.33}
\newcommand{\RTLingVLARSR}{43.20}
\newcommand{\RTLingVLALo}{36.60}
\newcommand{\RTLingVLAHi}{49.57}
\newcommand{\RTLingVLAMacro}{41.30}

\newcommand{\RTSmolSeed}{42.67}
\newcommand{\RTSmolRSR}{33.03}
\newcommand{\RTSmolLo}{27.66}
\newcommand{\RTSmolHi}{38.65}
\newcommand{\RTSmolMacro}{33.52}

\newcommand{\RTPiFiveSeed}{34.50}
\newcommand{\RTPiFiveRSR}{32.60}
\newcommand{\RTPiFiveLo}{27.48}
\newcommand{\RTPiFiveHi}{37.83}
\newcommand{\RTPiFiveMacro}{34.33}

\newcommand{\RTFastSeed}{81.00}
\newcommand{\RTFastRSR}{49.33}
\newcommand{\RTFastLo}{42.12}
\newcommand{\RTFastHi}{56.46}
\newcommand{\RTFastMacro}{55.62}

\newcommand{\RTLingVASeed}{83.75}
\newcommand{\RTLingVARSR}{59.50}
\newcommand{\RTLingVALo}{52.45}
\newcommand{\RTLingVAHi}{66.67}
\newcommand{\RTLingVAMacro}{53.19}

% LIBERO: state-bootstrap 95% CI and equal-weight Macro-9 point estimate.
\newcommand{\LBPiSeed}{85.17}
\newcommand{\LBPiRSR}{37.80}
\newcommand{\LBPiLo}{32.70}
\newcommand{\LBPiHi}{43.20}
\newcommand{\LBPiMacro}{36.79}

\newcommand{\LBPiFiveSeed}{94.17}
\newcommand{\LBPiFiveRSR}{64.40}
\newcommand{\LBPiFiveLo}{59.20}
\newcommand{\LBPiFiveHi}{69.50}
\newcommand{\LBPiFiveMacro}{67.71}

\newcommand{\LBBeingSeed}{91.67}
\newcommand{\LBBeingRSR}{46.80}
\newcommand{\LBBeingLo}{40.90}
\newcommand{\LBBeingHi}{52.70}
\newcommand{\LBBeingMacro}{48.68}

\newcommand{\LBUniSeed}{98.83}
\newcommand{\LBUniRSR}{48.00}
\newcommand{\LBUniLo}{41.60}
\newcommand{\LBUniHi}{54.60}
\newcommand{\LBUniMacro}{50.26}

\newcommand{\LBCosmosSeed}{97.83}
\newcommand{\LBCosmosRSR}{52.10}
\newcommand{\LBCosmosLo}{45.70}
\newcommand{\LBCosmosHi}{58.40}
\newcommand{\LBCosmosMacro}{50.56}

\newcommand{\LBFastSeed}{96.83}
\newcommand{\LBFastRSR}{54.00}
\newcommand{\LBFastLo}{47.80}
\newcommand{\LBFastHi}{60.10}
\newcommand{\LBFastMacro}{54.01}

% Rank movement and selected paired replay contrasts (percentage points).
\newcommand{\RTSpread}{26.90}
\newcommand{\LBSpread}{26.60}
\newcommand{\RTGapLo}{1.90}
\newcommand{\RTGapHi}{31.67}
\newcommand{\LBGapLo}{29.77}
\newcommand{\LBGapHi}{50.83}
\newcommand{\RTRankInv}{2}
\newcommand{\RTRho}{0.83}
\newcommand{\RTTau}{0.73}
\newcommand{\LBRankInv}{6}
\newcommand{\LBRho}{0.43}
\newcommand{\LBTau}{0.20}
\newcommand{\RTLVAminusX}{9.67}
\newcommand{\RTLVAminusXLo}{0.84}
\newcommand{\RTLVAminusXHi}{18.77}
\newcommand{\LBPiFiveMinusFast}{10.40}
\newcommand{\LBPiFiveMinusFastLo}{2.40}
\newcommand{\LBPiFiveMinusFastHi}{18.30}
\newcommand{\LBPiFiveMinusCosmos}{12.30}
\newcommand{\LBPiFiveMinusCosmosLo}{4.30}
\newcommand{\LBPiFiveMinusCosmosHi}{20.20}

% LIBERO clean-to-recovery reversal for UniFOLM versus pi_0.5 (points).
% The paired intervals resample the same 200 scenarios jointly.
\newcommand{\LBCleanUniMinusPiFive}{4.67}
\newcommand{\LBCleanUniMinusPiFiveLo}{2.33}
\newcommand{\LBCleanUniMinusPiFiveHi}{7.50}
\newcommand{\LBRecoveryPiFiveMinusUni}{16.40}
\newcommand{\LBRecoveryPiFiveMinusUniLo}{9.10}
\newcommand{\LBRecoveryPiFiveMinusUniHi}{23.80}

% LIBERO Finding 2: Being-H versus UniFOLM on the same 200 recovery
% scenarios. Diff = Being-H minus UniFOLM Overall RSR. ProfileDiff is
% mean_i |p_i,Being-H - p_i,UniFOLM| with a 10k scenario-bootstrap CI.
\newcommand{\LBBeingMinusUni}{-1.20}
\newcommand{\LBBeingMinusUniLo}{-9.00}
\newcommand{\LBBeingMinusUniHi}{6.70}
\newcommand{\LBProfileDiff}{40.40}
\newcommand{\LBProfileDiffLo}{35.00}
\newcommand{\LBProfileDiffHi}{46.00}
\newcommand{\LBBeingHigher}{60}
\newcommand{\LBUniHigher}{62}
\newcommand{\LBProfileTies}{78}

% Optimistic in-sample complementarity summaries.
\newcommand{\RTOracle}{95.63}
\newcommand{\RTOracleGain}{36.13}
\newcommand{\RTNoSuccess}{0}
\newcommand{\LBOracle}{91.80}
\newcommand{\LBOracleGain}{27.40}
\newcommand{\LBNoSuccess}{1}

% Observed-best alternative gains on natural source-policy rows (points).
\newcommand{\RTSourceGainLo}{40.28}
\newcommand{\RTSourceGainHi}{52.02}
\newcommand{\LBPiSourceGain}{38.75}
\newcommand{\LBPiSourceGainLo}{21.25}
\newcommand{\LBPiSourceGainHi}{67.50}
\newcommand{\LBPiFiveSourceGain}{61.67}
\newcommand{\LBPiFiveSourceGainLo}{43.33}
\newcommand{\LBPiFiveSourceGainHi}{85.00}

% High-support RoboTwin Stage--Type examples used in Finding 4.
\newcommand{\RTLingVAAppPose}{86.36}
\newcommand{\RTLingVAPlacePose}{21.74}
\newcommand{\RTFastPlacePose}{65.22}

% High-support LIBERO Stage--Type examples used in Finding 4.
\newcommand{\LBCosmosAppPose}{64.1}
\newcommand{\LBBeingPlacePose}{62.6}

% LIBERO construction-conditioned point-estimate leaders.
\newcommand{\LBNaturalBest}{70.48}
\newcommand{\LBPerturbedBest}{65.04}
\newcommand{\LBHumanBest}{70.32}

% RoboTwin downstream case studies; these are not primary same-state scores.
\newcommand{\RTMonitorX}{56.0}
\newcommand{\RTMonitorLing}{52.5}
\newcommand{\RTMonitorXTriggers}{152}
\newcommand{\RTMonitorLingTriggers}{166}
\newcommand{\RTRetreatX}{54.33}
\newcommand{\RTRetreatLing}{56.00}

% LIBERO exploratory recovery mechanisms from the latest aggregate report.
% These protocols have different starting-state/resource assumptions and are
% not entries in the primary direct-recovery leaderboard.
\newcommand{\LBFixedRewind}{74.0}
\newcommand{\LBMonitorRewind}{77.0}
\newcommand{\LBDeltaThirty}{71.2}
\newcommand{\LBMonitorDelta}{75.5}

\begin{abstract}
Robot-policy benchmarks increasingly cover diverse tasks and preset out-of-distribution conditions, but typically evaluate complete trajectories from predefined initial states. These evaluations often focus on the initialized scene and the final outcome, while paying less attention to the dynamic interaction process.   During closed-loop execution, actions and contacts can alter object relations and task progress, producing off-nominal intermediate states that need recovery.  Recovery requires a policy to infer how task progress has changed, correct the relevant relations, and continue the original goal.   We introduce \textbf{RoboRecover}, a benchmark for robot policy recovery under execution deviations.   RoboRecover selects deviation states from trajectories, reconstructs them by replaying action prefixes, and evaluates policies on the original task. RoboRecover contains 2,000 scenarios across RoboTwin and LIBERO, with 1,000 scenarios and a fixed 800/200 train/test split on each platform.  Results show that initial-state performance does not determine recovery performance and policies exhibit different recovery strengths across scenarios. Using its training split, RoboRecover further supports study on recovery interventions.  RoboRecover establishes recovery from execution-induced intermediate states as a distinct dimension of robot policy evaluation.
The project page is available on \GitHubLink.
\end{abstract}

\section{Introduction}

Generalist robot policies have made rapid progress across expanding manipulation suites, including CALVIN, LIBERO, RoboTwin, VLABench, and RoboCasa~\citep{mees2021calvin,liu2023libero, mu2025robotwin,zhang2024vlabench,nasiriany2024robocasa}. These robustness benchmarks vary in appearance, task composition, and other preset out-of-distribution conditions~\citep{pumacay2024colosseum, fei2025liberoplus,wang2024vlatest,kanwal2026fatevla}. However, these evaluations usually begin from predefined initial-states: they set the environment before policy execution. The final success/failure label summarizes the trajectory, but does not capture the dynamic interaction process.

\begin{figure}[t]
\centering
\includegraphics[width=\columnwidth]{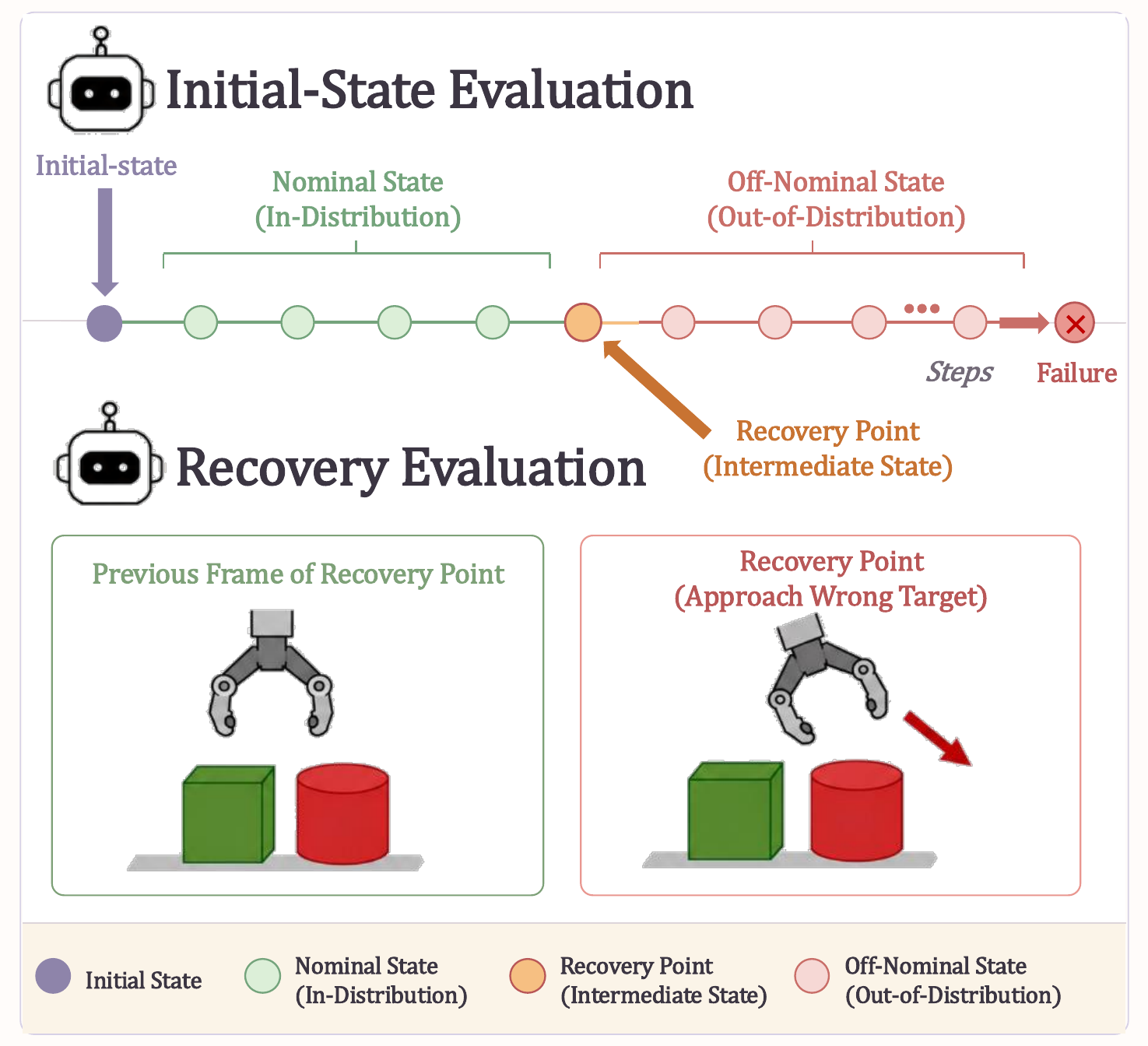}
\caption{\textbf{Initial-state evaluation vs. Recovery evaluation.} Recovery evaluation selects the recovery point at which an execution deviation occurs during interaction.}
\label{fig:protocol}
\end{figure}

Execution deviations emerge from policy interaction rather than difficult states defined before evaluation. Misdirected motion, unstable contact, or displaced objects can disrupt task progress or invalidate a completed subtask. Subsequent actions can compound the error and drive the system into an off-nominal intermediate state. Recovery requires the policy to reassess the current state, restore the relevant relations or contact, and continue the original task. Figure~\ref{fig:protocol} contrasts initial-state evaluation with recovery evaluation from execution-induced intermediate states.

These interaction-induced states raise issues that predefined initial-state evaluations cannot answer. We therefore study three questions. \textbf{Q1:} How well can current robot policies recover from intermediate states caused by execution deviations? \textbf{Q2:} Do policies show different recovery strengths across task stages, deviation types, and trajectory sources? \textbf{Q3:} Can interventions improve recovery from difficult intermediate states?

% 加上overview的图，表示initial-state和roborecover的思路
We introduce \textbf{RoboRecover}, a benchmark for evaluating robot policy recovery under execution deviations.   A recovery scenario is an intermediate task condition produced by deviations during dynamic robot–-environment interaction, while the original task remains meaningful and evaluable. Each scenario records the task reset, original instruction, recovery category, data source, and action prefix needed to reconstruct the recovery point. At evaluation time, the prefix is replayed and the candidate policy continues the original task from the reconstructed state.

RoboRecover contains \TotalRecords{} scenarios in RoboTwin and LIBERO, with \PlatformRecords{} scenarios and a fixed \TrainRecords{}/\TestRecords{} train/test split per platform. RoboTwin test set contains Natural scenarios collected from four source policies. LIBERO test set includes three sources: Natural scenarios, action-perturbed scenarios, and human-constructed scenarios. All scenarios undergo multi-stage filtering for data authenticity, difficulty, and reusability. 

We evaluate \NumPolicies{} policies per platform for each test scenario. Our results lead to some conclusions. Initial-state success does not determine recovery performance: on LIBERO, UnifoLM achieves \LBUniSeed\% initial-state success but only \LBUniRSR\% recovery, while $\pi_{0.5}$ reaches the best recovery rate of \LBPiFiveRSR\% despite a lower initial-state success rate of \LBPiFiveSeed\%. Capacity complementarity exists between policies: on RoboTwin, when source policies are evaluated on scenarios derived from themselves, they recover only 15.24--19.82\%, whereas other policies on the same source groups reach 58.06--70.27\%. Recovery performance varies across Recovery Stage and Deviation Type, so a single overall score cannot fully describe policy behavior.

Beyond evaluating fixed policies, we use the RoboRecover training split to study two intervention mechanisms. Reversion returns execution to an earlier state with less accumulated deviation. Correction learns from successful trajectories generated by alternative policies on scenarios where the base policy performs poorly. It briefly controls the robot from the recovery point before returning control to the unchanged base policy. These interventions demonstrate how RoboRecover can support the development and evaluation of recovery mechanisms.

Our contributions are as follows.
\begin{itemize}
    \item \textbf{Benchmark and protocol.} We introduce RoboRecover across RoboTwin and LIBERO, covering diverse construction sources and recovery categories. Unlike initial-state evaluation, its action-prefix replay protocol reconstructs the same recovery state for each policy and tests whether it can correct the deviation and continue the original task.

    \item \textbf{Recovery performance analysis.} We evaluate different embodied policies on each platform and show that high initial-state success does not guarantee effective recovery. On LIBERO, the policy achieves the highest initial-state success rate of \LBUniSeed\%, yet its recovery rate is only \LBUniRSR\%. We further analyze how recovery changes across task stages and deviation types.

    \item \textbf{Recovery mechanisms.} We propose two simple but effective methods: returning to an earlier, less-deviated state and applying a short correction learned from other policies before returning base-policy control. These results demonstrate the training split's utility and motivate further research on improving recovery capability.
\end{itemize}

\section{Related Work}

\subsection{General Manipulation Benchmarks}

CALVIN, LIBERO, RoboTwin, VLABench, and RoboCasa evaluate language-conditioned manipulation across diverse tasks, scenes, and embodiments~\citep{mees2021calvin,liu2023libero,mu2025robotwin,zhang2024vlabench,nasiriany2024robocasa}. SIMPLER, THE COLOSSEUM, and LIBERO-Plus extend this setting with visual, robot, and environmental variations~\citep{li2024simpler,pumacay2024colosseum,fei2025liberoplus}.  Although they broaden task diversity and preset Out-of-Distribution coverage, evaluation still begins from conditions specified before execution.  RoboRecover instead evaluates off-nominal intermediate states formed after robot actions and contacts have changed the scene and task progress.  It asks whether a policy can continue the original task from the consequences of prior interaction.

\subsection{Failure-Aware Recovery Benchmarks}

Recent work has extended robot-policy evaluation toward failure and recovery. RoboEval and recent SO-101 studies analyze failure stages, categories, and recovery behavior across complete trajectories~\citep{wang2025roboeval,yu2026so101}. FRBench with RePO-VLA evaluates recovery after predefined physical errors and uses recovery data to optimize a recovery-oriented policy~\citep{liufu2026repovla}. These works provide important failure analysis, but naturally occurring deviations are not their primary evaluation unit. RoboRecover addresses this gap by converting Natural deviations from policy--environment trajectories into recovery scenarios. It additionally includes action-perturbed and human-constructed scenarios, evaluates multiple policies on the same recovery points, and analyzes recovery across source policies, task stages, and deviation types. RoboRecover therefore provides a standardized evaluation of how current policies recover from errors that emerge during dynamic interaction, rather than focusing only on predefined error injection.
\section{Problem Formulation}
\label{sec:problem-formulation}
\noindent\textbf{Task and Initial-State Evaluation.}
We represent a task as $\mathcal{T}=(r,\ell,c,S)$, where $r$ is the environment reset, $\ell$ is the original instruction, $c$ is the success criterion, and $S$ is the maximum number of environment steps. In initial-state evaluation, a policy controls the trial from the predefined state produced by $r$. A trial succeeds if it satisfies $c$ within $S$ steps.

\noindent\textbf{Execution Deviation and Recoverability.}
A trajectory containing $T$ policy--environment interactions is $\tau=(o_0,a_0,o_1,a_1,\ldots,o_{T-1},a_{T-1},o_T)$, where $o_t$ is the observation at interaction $t$, $a_t$ is the action selected from $o_t$, and $o_{t+1}$ is the resulting observation after executing $a_t$. An \emph{execution deviation} occurs when previous actions or contacts disrupt nominal task progress and produce an \emph{off-nominal intermediate state}. This state is \emph{recoverable} if the original instruction and success criterion remain valid and the task could be completed by other policies.

\noindent\textbf{Recovery Point and Recovery State.}
A \emph{recovery point} is a selected time step $t$ at which a stable execution deviation is present and recovery evaluation begins. The corresponding environment state $x_t$ is the \emph{recovery state}, and $\mathbf{a}_{<t}=(a_0,\ldots,a_{t-1})$ is the \emph{action prefix} leading from the task reset to this state.

\noindent\textbf{Recovery Stage and Deviation Type.}
Each recovery state is described along two axes. The \emph{Recovery Stage} $z$ specifies the task phase that the policy must resume: Approach, Contact, Move, or Place. Approach requires approaching the correct object or region again. Contact requires restoring effective contact or grasp. Move requires resuming proper object transport or movement. Place requires moving to the target region after a valid operation.

The \emph{Deviation Type} $d$ specifies the observable error condition that must be corrected: Wrong Target, Wrong Pose, Stage Rollback, or Scene Change. Recovery Stage describes where task progress must resume, whereas Deviation Type describes what prevents normal interaction. Wrong Target indicates interaction with the wrong object or goal region. Wrong Pose indicates an unsuitable spatial relation between robot and object. Stage Rollback indicates lost contact, grasp, or previously completed task progress. Scene Change indicates that prior interaction has changed the scene and made the previous continuation unsuitable.

\noindent\textbf{Recovery Scenario and Evaluation.} Let $m$ denote the construction source, and let $x_t:=o_t$ denote the visual and robot-state observation at recovery point $t$. A \emph{recovery scenario} $\mathcal{R}=(\mathcal{T},\mathbf{a}_{<t},m,z,d)$ combines the original task $\mathcal{T}$, action prefix $\mathbf{a}_{<t}$, construction source $m$, Recovery Stage $z$, and Deviation Type $d$ as a fixed evaluation unit. During evaluation, the environment is reset using $r$, and $\mathbf{a}_{<t}$ is replayed through the simulator to reach the recovery point and produce $x_t$. The candidate policy initializes its observation/action history at $t$, receives $x_t$ and the original instruction $\ell$, and begins recovery. It receives up to $S$ new environment interactions, while replay actions do not count toward this limit.

\section{RoboRecover}
\subsection{Benchmark Overview}
\begin{figure*}[t]
\centering
\includegraphics[width=\textwidth]{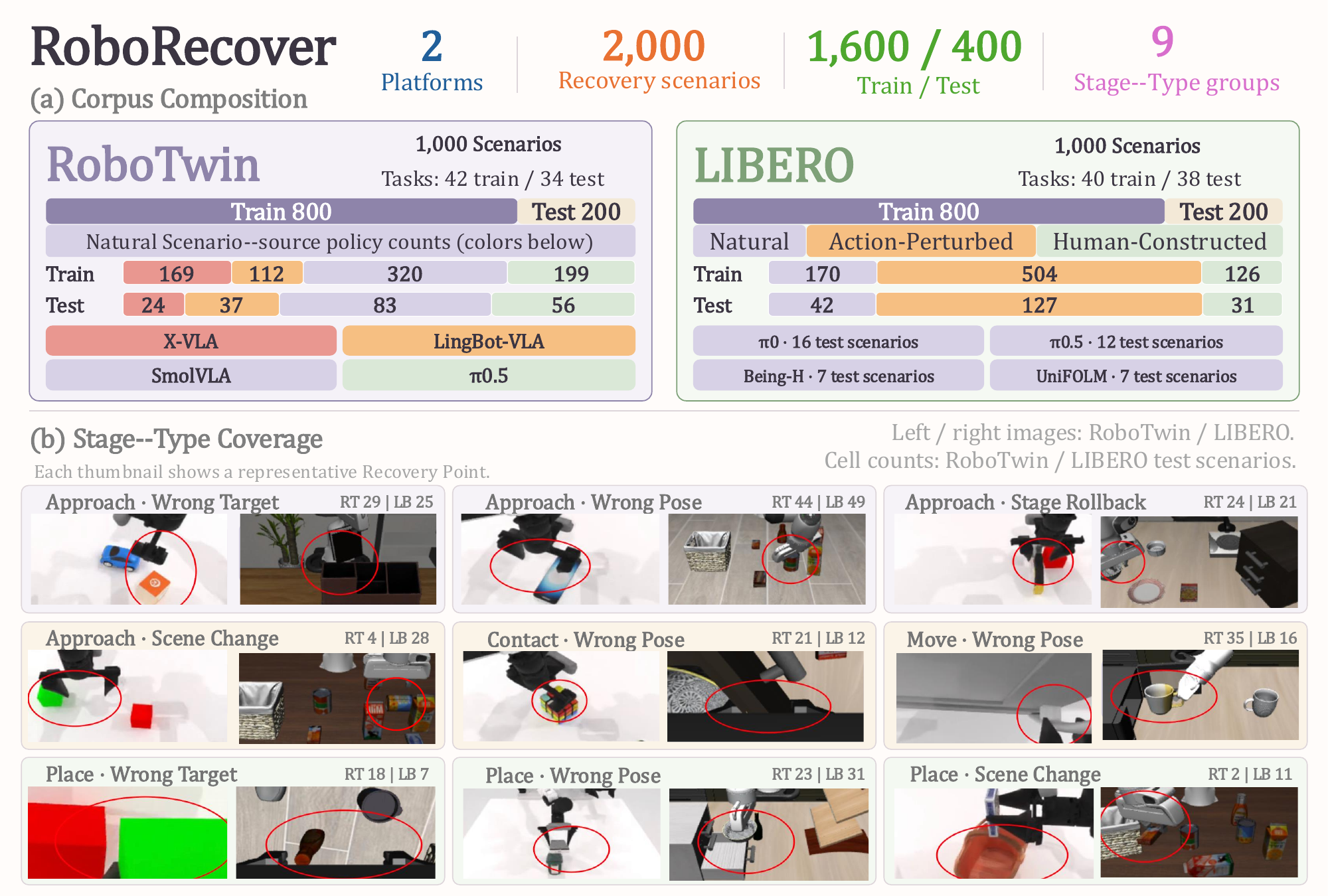}
\caption{\textbf{RoboRecover benchmark composition and coverage.} The upper panels summarize platform, split, task, construction source, and source policies. The lower matrix shows one representative recovery point for each of the nine Stage--Type groups; cell counts report RoboTwin/LIBERO test scenarios.}
\label{fig:bench}
\end{figure*}

RoboRecover contains \TotalRecords{} recovery scenarios across RoboTwin and LIBERO, with \PlatformRecords{} scenarios per platform and fixed \TrainRecords{}/\TestRecords{} train/test splits. RoboTwin contains scenarios from 42 training tasks and \RTTestTasks{} test tasks, while LIBERO contains scenarios from 40 training tasks and \LBTestTasks{} test tasks. The training splits support recovery method development and all primary policy comparisons use the fixed test splits. Figure~\ref{fig:bench} summarizes the data composition and shows the representative recovery points from Stage--Type occupied group.

RoboTwin consists of Natural scenarios collected from X-VLA, LingBot-VLA, SmolVLA, and $\pi_{0.5}$~\citep{zheng2025xvla,wu2026lingbotvla,shukor2025smolvla,physicalintelligence2025pi05}. LIBERO combines Natural scenarios collected from $\pi_{0}$, $\pi_{0.5}$, Being-H0.5, UnifoLM\citep{black2025pi0,physicalintelligence2025pi05,luo2026beingh05,unitree2026unifolmvla}, additionally includes action-perturbed and human-constructed scenarios. Construction source describes how a scenario was produced.

The benchmark covers four Recovery Stages: Approach, Contact, Move, Place; and four Deviation Types: Wrong Target, Wrong Pose, Stage Rollback, and Scene Change, as defined in the problem formulation. Nine Stage–Type combinations are present in both fixed test sets.

% \begin{figure*}[t]
% \centering
% \includegraphics[width=\textwidth]{figures/f2.pdf}
% \caption{\textbf{Recovery scenario construction pipeline.} Candidate trajectories from three sources pass recovery point selection, Stage--Type annotation, difficulty screening, replay-based quality control, final review, and train/test assignment.}
% \label{fig:construction-pipeline}
% \end{figure*}
  
\subsection{Scenario Sources and Construction}

We construct candidate trajectories from three sources: \emph{Natural} policy trajectories, \emph{Action-Perturbed} trajectories, and \emph{Human-Constructed} trajectories.  In all three sources, deviations arise through robot--environment interactions dynamically rather than direct simulator state editing. 
% The candidates then enter the common selection and quality-control pipeline shown in Figure~\ref{fig:construction-pipeline}.

\noindent\textbf{Natural Scenarios.} Natural candidates are extracted from failed trajectories generated by multiple source policies. We exclude erratic or out-of-region motion, irreversible task damage, and all states where the robot cannot continue to interact correctly.  Each accepted Natural scenario records of its source policy information.   

\noindent\textbf{Action-Perturbed Scenarios.} Action-perturbed candidates are also generated from the interaction of policies. However, the perturbation is accepted only in the first 150 interaction steps, and the policy inference after 150 steps is not affected. During the first 150 interaction steps, we independently perturb all three translational coordinates of each action between policy prediction and environment execution:
\begin{equation}
  \begin{aligned}
    \epsilon_d &\sim \operatorname{Uniform}(-1.0,1.0),
      && d\in\{x,y,z\},\\
    a'_d &= a_d+\epsilon_d.
  \end{aligned}
\end{equation}
The perturbation is defined in normalized robot action space. If the perturbed action is not reachable by inverse kinematics, we execute the original action. After perturbation, the robot usually falls into an off-nominal state that is difficult to execute.

\noindent\textbf{Human-Constructed Scenarios.} Human-constructed scenarios are added on LIBERO after the Natural and Action-Perturbed candidates have been fully screened. They supplement task-relevant states that remain rare in the filtered data because LIBERO policies have high success rates and seldom produce such deviations, such as \textit{Place--Scene Change}. These states are recoverable but remain challenging for most policies. Operators use a SpaceMouse through the simulator interface to create deviation and then complete the original task to verify recoverability. All the trajectories are produced through remote control method, then undergo the same quality-control process.

\subsection{Stage--Type and Recovery Point Annotation}

Recovery points and Stage--Type labels are determined in the same annotation process. Three annotators review each complete trajectory using a shared guide and identify candidate points where a deviation causes the subsequent part of the trajectory to enter an off-nominal state. As a result, the recovery point is usually neither the first unusual frame nor the last frame of a failed episode. For Natural and action-perturbed trajectories, a candidate point is discarded if later policy actions restore nominal progress, indicating that the deviation is temporary. For human-constructed trajectories, the operator's deliberate correction instead confirms that the selected state is recoverable. Annotators select the final stable deviation point, which subsequent steps stuck into a sustained failure.

Recovery point identification and Stage--Type labeling are performed simultaneously within a single annotation pass. As annotators examine each trajectory, they mark candidate recovery points and, at the same time, assign the corresponding Recovery Stage and Deviation Type based on the observed state at that point. Labels are determined by majority vote across annotators, and scenarios without consensus are removed. We do not force equal category sizes: the benchmark largely preserves the distribution observed in Natural policy rollouts, while action-perturbed and human-constructed scenarios supplement uncommon but recoverable cases. 

\subsection{Scenario Filtering}
%更改一下标题，主要是难度过滤，先总再分来介绍
% \noindent\textbf{Policy Difficulty Filtering.} Each candidate is evaluated in five independent trials by four VLA policies, giving 20 trials in total. RoboTwin uses X-VLA, LingBot-VLA, SmolVLA, and $\pi_{0.5}$, while LIBERO uses $\pi_0$, $\pi_{0.5}$, Being-H0.5, and UnifoLM. We retain candidates with 2--18 successful trials, corresponding to a recovery rate between 10\% and 90\%. This screening removes nearly trivial and nearly unsolvable scenarios and is applied to all sources before the train/test split.
\noindent\textbf{Policy Difficulty Filtering.} We filter candidates by keeping only scenarios of intermediate difficulty, as measured by how often a set of reference policies can successfully recover from them. Each candidate is evaluated in five independent trials by four VLA policies, giving 20 trials in total. RoboTwin uses X-VLA, LingBot-VLA, SmolVLA, and $\pi_{0.5}$, while LIBERO uses $\pi_0$, $\pi_{0.5}$, Being-H0.5, and UnifoLM. We retain candidates with 2--18 successful trials, corresponding to a recovery rate between 10\% and 90\%. This screening removes nearly trivial and nearly unsolvable scenarios and is applied to all sources before the train/test split.

% 可以先重播验证，再精细过滤
 % We retain trajectories containing intermediate deviations from which the original task remains observable and recoverable.
\noindent\textbf{Replay-Based Data Checking.} After difficulty filtering, each candidate undergoes replay validation. Some difficult states created during construction are sensitive to replay errors, which may cause the reconstructed state to differ from the original and lead the policy to infer from an unintended condition. We therefore inspect three inference trajectories from the preceding policy-filtering trials and verify that replay reconstructs the same recovery condition. Candidates with inconsistent states, simulator crashes, or invalid actions are removed.

To provide an objective consistency criterion, we compare both simulator states and rendered observations between the original and replayed trajectories.    We measure robot and object state deviations directly from the simulator, requiring the end-effector position deviation to be below 2 mm and the object position deviation to be below 2 mm.    We additionally compute the Structural Similarity Index Measure (SSIM) between the original and replayed RGB observations over the final five frames before the recovery point, and retain scenarios with an average SSIM score above 0.99.    Manual inspection is performed only as a final sanity check to remove rare cases with physically inconsistent states that are not captured by numerical criteria.

% Each retained candidate is reconstructed from its task reset by replaying the recorded action prefix. Reviewers compare the reconstructed and original trajectories at the recovery point and during the first policy actions. A candidate is retained only if:(1) replay finishes without an environment crash, interface stall, or invalid action; (2) the robot, relevant objects, and task progress correspond to the original trajectory. 

% \noindent\textbf{Fixed-Version Review.} The remaining candidates are replayed using fixed simulator versions, task assets, and environment configurations. A final manual review checks scenario usability, replay stability, and annotation consistency before the train/test split is finalized.

% \subsection{Benchmark Composition and Data Usage}
% \label{sec:data}

% RoboRecover contains \TotalRecords{} scenarios, evenly divided between RoboTwin and LIBERO. Each platform uses a fixed \TrainRecords{}/\TestRecords{} train/test split and evaluates \NumPolicies{} policies. The RoboTwin test set covers \RTTestTasks{} tasks and contains Natural scenarios, while the LIBERO test set covers \LBTestTasks{} tasks and additionally includes Action-Perturbed and Human-Constructed scenarios. Primary policy comparisons use the held-out test sets. The training scenarios support recovery-monitor learning, and the LIBERO training split is also used to train the Delta model. Detailed task composition and category distributions are provided in the supplement.
\section{Experiments}
% 前面先定义四个问题，1. 体现模型很难，解决不了，2. cross-policy，是不是一个模型rollout构造的，本身构造的很好或其他构造的更好，3. recovery能力是不是在不同阶段不同 4. 是不是能定义一个方法来解决
Our experiments are organized around 4 questions:

\textbf{\textit{Q1:} }Can initial-states success predict recovery performance in recovery states?

\textbf{\textit{Q2:} }Does a policy’s recovery performance stay consistent across stages and deviation types?

\textbf{\textit{Q3:} }For a scenario generated by one policy, is that policy also the best at recovery, or can another policy recover better?

\textbf{\textit{Q4:} }Are there any methods that can improve the recovery capability of policy?

\subsection{Experimental Setup}
\label{sec:exp-scope}

We evaluate six policies on each platform, selected to cover VLA and world-action architectures with different pretraining and action-generation designs. RoboTwin includes X-VLA, LingBot-VLA, SmolVLA, $\pi_{0.5}$, Fast-WAM, and LingBot-VA~\citep{yuan2026fastwam,li2026lingbotva}. LIBERO includes $\pi_0$, $\pi_{0.5}$, Being-H0.5, UnifoLM, Cosmos-Policy, and Fast-WAM~\citep{kim2026cosmospolicy,yuan2026fastwam}.

% 在这部分定义的时候对比initial和recover的评测区别，对比了什么
Model selection follows two principles. First, we choose distinct policies trained across multiple tasks within each platform. Second, we prioritize models with publicly available Hugging Face checkpoints, high community adoption, and strong reported performance, ensuring that the evaluated policies are both competitive and accessible. 

% \subsection{Evaluation Metrics}
% For scenario $i$, policy $p$, and recovery trial $k$, let $y_{i,p,k}\in\{0,1\}$ indicate whether task success, and let $K_{i,p}$ be the number of recovery trials. We compute the scenario-level recovery rate and Overall Recovery Success Rate (RSR) as: 
% \begin{equation}
% \hat p_{i,p}=\frac{1}{K_{i,p}}\sum_{k=1}^{K_{i,p}}y_{i,p,k},\qquad\mathrm{RSR}p=\frac{1}{N}\sum{i=1}^{N}\hat p_{i,p}.
% \end{equation}
% Overall RSR is the primary metric and gives every scenario equal weight.

% \emph{Initial-State Success Rate} measures full-task success from matched predefined initial-states and provides a reference for recovery performance.

% For Stage--Type analysis, we average scenario-level recovery rates within each of the nine nonempty groups. \emph{Macro-9 RSR} gives these groups equal weight and serves as a category-balanced secondary metric.

% For Natural scenarios, we group scenarios by the source policy whose trajectory produced the deviation and evaluate every candidate policy on each group. This analysis compares a source policy's recovery from its own deviations with the recovery of other policies.

% 主要分析initial state，放LIBERO的就好，补一小段说表面表现的好，但其实遇到action ood就失效
\subsection{Initial-State and Recovery Performance}

\vpara{Setup and Metrics.}
We evaluate each policy on the same task configurations under two starting conditions. Initial-state evaluation is reset before policy inference, whereas recovery evaluation begins from the reconstructed recovery points after replaying action prefix. Both settings use the original instruction and success criterion. Initial-State Success Rate is the fraction of trials completed from the predefined initial state.

\begin{figure*}[t]
\centering
\includegraphics[width=\textwidth]{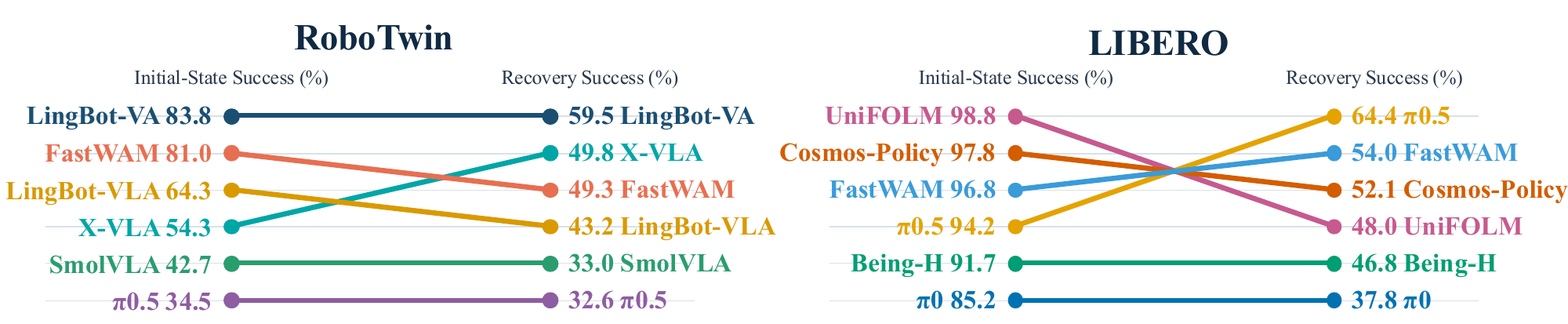}
\caption{\textbf{Policy performance changes from initial-state to recovery evaluation.} Lines connect each policy's rank under matched initial-state and recovery evaluation.}
\label{fig:rank-transition}
\end{figure*}

For recovery scenario $i$, policy $p$, and trial $k$, let $y_{i,p,k}\in\{0,1\}$ indicate whether the original task is completed, and let $K_{i,p}$ denote the number of trials. We compute the scenario-level recovery rate and Overall Recovery Success Rate (RSR) as
\begin{equation}
\hat p_{i,p}
=\frac{1}{K_{i,p}}\sum_{k=1}^{K_{i,p}}y_{i,p,k},
\qquad
\mathrm{RSR}_p
=\frac{1}{N}\sum_{i=1}^{N}\hat p_{i,p}.
\end{equation}
Overall RSR is the primary recovery metric and gives every scenario equal weight. Figure~\ref{fig:rank-transition} compares the evaluation difference.

\vpara{Findings: High Initial-State Success Does Not Guarantee Strong Recovery.}
Every policy performs worse under recovery evaluation, but the magnitude of the decrease varies substantially. On LIBERO, all six policies achieve at least \LBPiSeed\% success from predefined initial states, yet the best recovery result is only \LBPiFiveRSR\%, achieved by $\pi_{0.5}$. Because the task, instruction, and success criterion remain unchanged, this gap mainly reflects the robot--scene states produced by prior interaction. It indicates a trajectory-level distribution shift: policies that perform well along nominal trajectories may still struggle with uncommon intermediate configurations, altered object relations, and partially invalidated task progress.

For example, UnifoLM decreases from \LBUniSeed\% initial-state success to \LBUniRSR\% recovery, whereas $\pi_{0.5}$ decreases from \LBPiFiveSeed\% to \LBPiFiveRSR\%. Their different performance drops show that strong initial-state results can conceal different sensitivities to execution-induced states. Both policies use pretraining, so the presence of pretraining alone does not explain this difference. $\pi_{0.5}$ combines heterogeneous robot and web data with high-level subtask prediction and flow-matched action chunks, whereas UnifoLM uses continued VLM and robot pretraining with a DiT-based diffusion controller. This comparison suggests that diverse training data and explicit task decomposition may help a policy reassess partial task progress after a deviation. Fast-WAM similarly decreases from \LBFastSeed\% to \LBFastRSR\% without embodied pretraining, suggesting that video representation learning and co-training may provide useful information for interpreting changes across time.

\subsection{Stage--Type Recovery Differences}
\begin{figure*}[t]
\centering
\includegraphics[width=\textwidth]{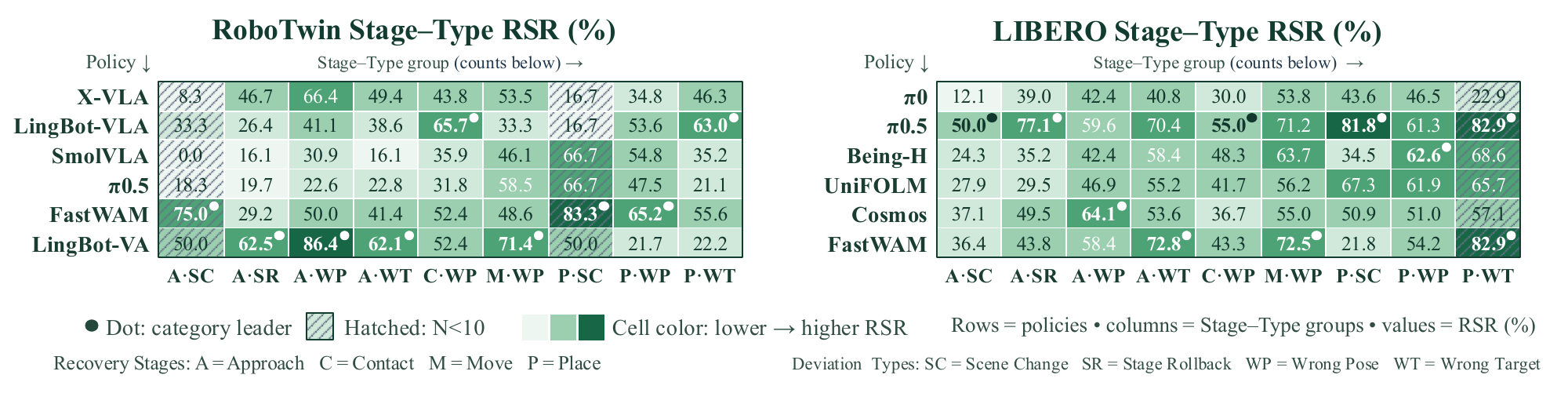}
\caption{\textbf{Recovery rates across Stage--Type groups.}
Cells include scenario counts, and sparse groups are visually de-emphasized. Differences between groups can be hidden by Overall RSR. A·SC: Approach-Scene Change.         }
\label{fig:recovery-profiles}
\end{figure*}

\vpara{Setup and Metrics.} We divide the fixed test scenarios into the nine groups defined by Recovery Stage and Deviation Type. For each policy, the recovery rate of a Stage--Type group is obtained by averaging the scenario-level recovery rates within that group. Macro-9 RSR then averages the nine group rates with equal weight, whereas Overall RSR gives greater influence to groups containing more scenarios. Macro-9 RSR is used as a category-balanced diagnostic because several groups contain relatively few scenarios. Figure~\ref{fig:recovery-profiles} reports the group-level results and scenario counts.

% \noindent\textbf{Recovery Varies with Recovery Stage and Deviation Type.} A policy that performs well in one Stage--Type setting may fail in another, indicating non-uniform recovery behavior across the trajectory. Figure~\ref{fig:recovery-profiles} shows performance across all Stage--Type groups.

% On RoboTwin, LingBot-VA achieves \RTLingVAAppPose\% on Approach--Wrong Pose but drops to \RTLingVAPlacePose\% on Place--Wrong Pose, where Fast-WAM performs best with \RTFastPlacePose\%. On LIBERO, Cosmos Policy reaches \LBCosmosAppPose\% on Approach--Wrong Pose, while Being-H0.5 performs best on Place--Wrong Pose with \LBBeingPlacePose\%. These results show that even the same deviation type can require different recovery strategies depending on the task stage. Thus, recovery should be viewed as a structured performance profile rather than a single score.

\vpara{Findings 1: The Same Deviation Type Requires Different Recovery Across Task Stages.} Recovery performance varies substantially across Recovery Stages and Deviation Types. On RoboTwin, LingBot-VA achieves \RTLingVAAppPose\% on Approach--Wrong Pose but only \RTLingVAPlacePose\% on Place--Wrong Pose, where Fast-WAM reaches \RTFastPlacePose\%. On LIBERO, Cosmos Policy achieves \LBCosmosAppPose\% on Approach--Wrong Pose, whereas Being-H0.5 reaches \LBBeingPlacePose\% on Place--Wrong Pose.

Although both groups contain a Wrong Pose deviation, they require different corrective behavior. Approach--Wrong Pose mainly requires the policy to re-localize the relevant object or target and produce an approach trajectory. Place--Wrong Pose occurs after more interaction has already taken place and requires the policy to correct the object relation while grasping the object. This helps explain why strength on one Stage--Type group does not necessarily transfer to another.

\vpara{Findings 2: Stage--Type Weighting Changes Aggregate Conclusions.} The difference also affects aggregate evaluation. On RoboTwin, LingBot-VA has the highest Overall RSR at \RTLingVARSR\%, while Fast-WAM reaches the highest Macro-9 RSR at \RTFastMacro\%. This change shows that conclusions can depend on how frequently different recovery situations occur in the test set. On LIBERO, $\pi_{0.5}$ performs best under both metrics, indicating that its advantage is less sensitive to Stage--Type weighting. Recovery should therefore be examined across Stage--Type groups rather than described only by a single Overall RSR.

\subsection{Cross-Policy Recovery Differences}
\begin{figure*}[t]
\centering
\includegraphics[width=\textwidth]{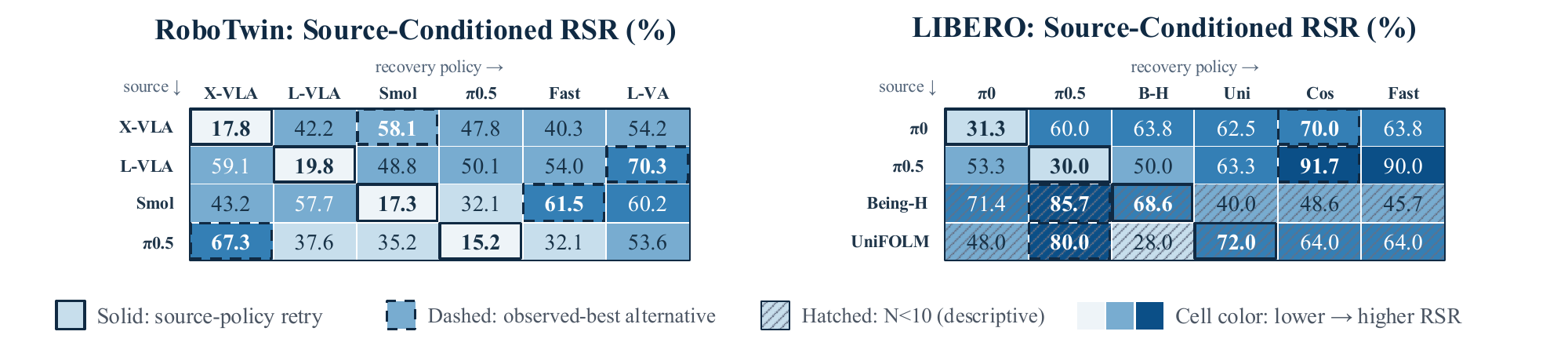}
\caption{\textbf{Cross-policy recovery on Natural deviations.} Rows identify source policies, and columns show the evaluated recovery policies. Solid outlines mark source-policy re-evaluation; dashed outlines mark the observed-best alternative.}
\label{fig:source-recovery}
\end{figure*}

\vpara{Setup.}This experiment uses Natural scenarios, which retain the policy whose rollout produced each deviation. We group scenarios by their source policy and evaluate all six candidate recovery policies. Diagonal comparisons measure source-policy retry, while off-diagonal comparisons measure cross-policy recovery. Figure~\ref{fig:source-recovery} reports these results. To measure policy complementarity beyond source groups, we also construct a retrospective oracle that selects the highest observed recovery rate among the six policies for each test scenario. 

\vpara{Findings: Cross-Policy Recovery Can Overcome Policy-Specific Failures.} On RoboTwin, the source policy is not the strongest recoverer in any of the four source groups. Source-policy recovery ranges from 15.24--19.82\%, whereas the observed-best alternative reaches 58.06--70.27\%. LIBERO shows a similar pattern in its two best-supported source groups. This suggests that many deviations are not inherently unrecoverable: the source policy remains unsuccessful, while another policy can produce a suitable trajectory from the same scenario.

The retrospective oracle reaches \RTOracle\% on RoboTwin and \LBOracle\% on LIBERO, exceeding the best fixed policy. This gap shows that the evaluated policies recover substantially different sets of scenarios. It additionally motivates the Correction model in Section~\ref{sec:intervention}: when a fixed base policy fails but another policy succeeds, the successful cross-policy trajectory can provide a corrective demonstration for learning recovery behavior.
% \noindent\textbf{Recovery Performance Depends on Both Source and Recovery Policies.} Recovery ability is not an intrinsic property of a single policy; it depends jointly on the policy that caused the failure and the one attempting recovery. Since all policies are evaluated on the same reconstructed states in Natural scenarios, we can directly compare cross-policy behavior. Figure~\ref{fig:source-recovery} reports recovery performance grouped by source policy.

% On RoboTwin, a policy is rarely its own best recovery option. Its self-recovery ranges from 15.24--19.82\%, while the best alternative policy reaches 58.06--70.27\% depending on the source group. Moreover, the best recovery policy varies across source policies, indicating that different failure origins require different strategies. On LIBERO, Cosmos Policy improves recovery by \LBPiSourceGain{} points in $\pi_0$-source scenarios and by \LBPiFiveSourceGain{} points in $\pi_{0.5}$-source scenarios. Additional results are in the supplement. Overall, recoverability is highly dependent on the interaction between source and recovery policies.

% \noindent\textbf{Policies Exhibit Complementary Recovery Strengths.} The observed variation suggests that different policies specialize in different recovery regimes. To quantify this complementarity, we construct an oracle that selects the best observed recovery policy per scenario. This oracle achieves \RTOracle\% on RoboTwin and \LBOracle\% on LIBERO, outperforming the best single policy by \RTOracleGain{} and \LBOracleGain{} points, respectively. Although this oracle is not directly deployable because it selects the best-performing policy using evaluation-time outcome information (i.e., it assumes we already know which recovery attempt succeeds), it reveals substantial headroom and highlights the benefit of combining multiple policies rather than relying on a single one. This also provides ideas for the construction of the Delta Model mentioned in Section~\ref{sec:intervention}

\subsection{Benchmark-Guided Recovery Interventions}
\label{sec:intervention}
% Recovery requires a policy to correct a difficult intermediate state without additional support. However, the previous experiments show that many scenarios failed by one policy can still be recovered by another. RoboRecover training split can additionally support the development of recovery intervention mechanisms. We consider two perspectives, one is to keep the policy capability constant and reduce the recovery difficulty, and the other is to keep the recovery difficulty constant and briefly boost the policy near the recovery point.

% Using only the 800 LIBERO training scenarios, we construct two interventions around a frozen $\pi_{0.5}$-based policy. \textit{Reversion} acts on the recovery state by returning to an earlier point with less accumulated deviation. \textit{Correction} model acts on the controller by temporarily handing control to a learned correction policy. % A \textit{recovery monitor} can optionally decide when either intervention is activated. %All results use the fixed 200-scenario test split, five trials per scenario, and the original task-success predicate.
The cross-policy results show that states that are difficult for one policy can remain recoverable by another. We therefore use the 800 LIBERO training scenarios to develop two interventions around the $\pi_{0.5}$. \emph{Reversion} reduces accumulated deviation by starting 30 steps before the recovery point. \emph{Correction} leaves the recovery point unchanged but temporarily transfers control to a learned correction policy for 30 steps.

\vpara{Setup.} Correction is trained from complementary cross-policy behavior. We select training scenarios where direct $\pi_{0.5}$ recovery is below 40\% and at least one alternative policy exceeds 60\%. The alternative policies are evaluated four times, and their successful recovery trajectories are retained as correction demonstrations. These trajectories train a separate $\pi_{0.5}$-based controller with the same observation and action interface. After 30 steps, control returns to the original $\pi_{0.5}$.

We train a recovery monitor on the LIBERO training split. Four VLA screening policies each perform one trajectory on every training scenario, producing 3,200 trajectories from the 800 scenarios. For each trajectory, frames strictly before the recovery point are labeled 0, while the recovery-point frame and all subsequent frames are labeled 1. We fine-tune Qwen3-VL-4B~\citep{bai2025qwen3vl} as a binary classifier. During evaluation, the first positive prediction activates Reversion or Correction, after which monitoring stops.

All methods are evaluated on the fixed 200-scenario test split using five trials per scenario. These intervention results are separate from the primary same-state policy leaderboard because Reversion changes the starting state and Correction adds an external controller.

\begin{table}[t]
\centering
\small
\setlength{\tabcolsep}{1mm}
\renewcommand{\arraystretch}{1.05}
\begin{tabular*}{\columnwidth}
{@{\extracolsep{\fill}}lp{0.49\columnwidth}r@{}}
\toprule
Setting & Recovery procedure & RSR (\%) \\
\midrule
Direct $\pi_{0.5}$
& Run $\pi_{0.5}$ from the recovery point
& 64.4 \\

Reversion
& Start 30 steps before the recovery point; run $\pi_{0.5}$
& 74.0 \\

Monitor+Reversion
& Run $\pi_{0.5}$; at the first trigger, rollback 30 steps and resume
& 77.0 \\

Correction
& Run Correction for 30 steps from the recovery point; resume $\pi_{0.5}$
& 71.2 \\

Monitor+Correction
& Run $\pi_{0.5}$; at the first trigger, apply Correction for 30 steps
& 75.5 \\
\bottomrule
\end{tabular*}

\normalsize
\caption{\textbf{LIBERO recovery-intervention results. }Reversion changes the startingstate, whereas Correction adds a 30-step learned controller. Monitor-basedsettings activate the corresponding intervention selectively.}
\label{tab:mechanism-utility}
\end{table}

\vpara{Different Interventions Both Improve Recovery.}Direct $\pi_{0.5}$ recovery reaches 64.4\%. Reversion increases recovery to 74.0\%, showing that reducing the deviation accumulated before the recovery point makes recovery easier for the unchanged base policy. Correction reaches 71.2\%, showing that successful cross-policy trajectories can be transferred into short corrective behavior. The two interventions therefore improve recovery through different routes: Reversion changes the state presented to the policy, whereas Correction temporarily strengthens the controller at the original recovery point.

\vpara{Monitor Further Improves Intervention Effectiveness.} Adding a monitor improves both intervention strategies: Monitor+Reversion reaches 77.0\%, and Monitor+Correction reaches 75.5\%. In both cases, performance is higher than the corresponding interventions without monitoring, indicating that the monitor helps identify states where interventions are most beneficial. This suggests that beyond simply detecting likely failures of direct recovery, the monitor also provides useful guidance for when each intervention should be applied. These results highlight that RoboRecover benefits from combining learned intervention mechanisms with state-aware triggering, and that the training split can be used to develop such mechanisms while the fixed test split remains the final evaluation target.

\section{Conclusion}

\vpara{Summary.}
We introduce RoboRecover, a benchmark for evaluating robot policies from replayable intermediate states created during execution. Its action-prefix replay protocol enables different policies to be evaluated from the same reconstructed recovery state, separating recovery capability from conventional initial-state task completion. Across RoboTwin and LIBERO, policy performance changes substantially between initial-state and recovery evaluation, while the strongest recovery policy varies across deviation sources, types, and task stages. These results show that high initial-state success does not guarantee that a policy can interpret an off-nominal state, repair disrupted task progress, and continue the original task. Our intervention experiments further demonstrate that recovery can be improved either by returning execution to an earlier state with less accumulated deviation or by applying a short correction learned from successful cross-policy trajectories. RoboRecover therefore establishes recovery as an explicit dimension of robot policy evaluation and demonstrates how its training split can support targeted recovery mechanisms.

\vpara{Future Work.}
RoboRecover currently covers two simulation platforms, and extending it to additional task suites, robot embodiments, and real-world systems would broaden its scope. The recovery mechanisms studied here provide an initial demonstration of benchmark-guided method development rather than an exhaustive exploration of recovery strategies. Future work will evaluate broader policy families and develop advanced recovery mechanisms to detect execution deviations and generate corrective behaviors across diverse tasks and environments.
\bibliography{refs/references}

% The appendix is in this same arXiv PDF rather than a separate supplement,
% and begins on a fresh page.
\clearpage
\appendix
% !TeX root = ../Appendix2027.tex
\section{Benchmark Construction and Composition}
The main paper describes the three construction sources, the Stage--Type taxonomy, and the recovery-point selection criteria. This section reports additional annotation, validation, and corpus statistics.
\subsection{Annotation and Recovery-Point Selection}
% 三名标注者、同时选择 recovery point 和 Stage--Type
% 两项一致率约 92.17% 和 82.82%；九个非空 Stage--Type 单元格
Three annotators independently review each complete trajectory.  During the same annotation pass, they select the stable recovery point and assign its Recovery Stage and Deviation Type.  Labels are resolved by majority vote, and scenarios without a stable majority are removed.  Before majority resolution, the agreement rates for Recovery Stage and Deviation Type are approximately 92.17\% and 82.82\%, respectively.

The fixed test sets contain the same nine nonempty Stage--Type combinations, but their frequencies are not balanced. Table~\ref{tab:app-stage-type-counts} reports the support of each group.

\noindent\begin{minipage}{\columnwidth}
\centering
\small
\setlength{\tabcolsep}{4.2pt}
\renewcommand{\arraystretch}{1.02}
\begin{tabular*}{\columnwidth}
{@{\extracolsep{\fill}}lrr@{}}
\toprule
Stage--Type & RoboTwin & LIBERO \\
\midrule
Approach--Wrong Target   & 29 & 25 \\
Approach--Wrong Pose     & 44 & 49 \\
Approach--Stage Rollback & 24 & 21 \\
Approach--Scene Change   &  4 & 28 \\
Contact--Wrong Pose      & 21 & 12 \\
Move--Wrong Pose         & 35 & 16 \\
Place--Wrong Target      & 18 &  7 \\
Place--Wrong Pose        & 23 & 31 \\
Place--Scene Change      &  2 & 11 \\
\midrule
Total                     & 200 & 200 \\
\bottomrule
\end{tabular*}
\captionof{table}{\textbf{Numbers of fixed test scenarios in the nine occupied
Stage--Type groups.}}
\label{tab:app-stage-type-counts}
\end{minipage}

Figure~\ref{fig:app-stage-type-examples} pairs one RoboTwin and one LIBERO
example for every occupied Stage--Type group. Each image shows the observation
at the selected recovery point, before recovery inference begins.

\begin{figure*}[t]
\centering
{\small
\setlength{\tabcolsep}{4pt}
\newcommand{\stageexampleimage}[1]{%
  \IfFileExists{#1}{%
    \includegraphics[width=0.125\textwidth,height=0.56in,
      keepaspectratio]{#1}}{%
    \fbox{\parbox[c][0.50in][c]{0.115\textwidth}{%
      \centering Image pending}}}}
\newcommand{\stageexamplecard}[4]{%
  \parbox[t][1.20in][t]{0.305\textwidth}{%
    \centering\textbf{#1}\par
    \begin{tabular}{@{}cc@{}}
      \stageexampleimage{#2} & \stageexampleimage{#3} \\
      RoboTwin & LIBERO
    \end{tabular}\par
    \raggedright #4\par}}
\begin{tabular}{@{}ccc@{}}
\toprule
\stageexamplecard{Approach--Wrong Target}{figures/stage_type_examples/robotwin_approach_wrong_target.png}{figures/stage_type_examples/libero_approach_wrong_target.png}{Redirect from an irrelevant object or region and approach the correct task target.} &
\stageexamplecard{Approach--Wrong Pose}{figures/stage_type_examples/robotwin_approach_wrong_pose.png}{figures/stage_type_examples/libero_approach_wrong_pose.png}{Correct the robot--object relation before attempting the intended interaction.} &
\stageexamplecard{Approach--Stage Rollback}{figures/stage_type_examples/robotwin_approach_stage_rollback.png}{figures/stage_type_examples/libero_approach_stage_rollback.png}{Recover lost task progress and return to a valid approach phase.} \\
\stageexamplecard{Approach--Scene Change}{figures/stage_type_examples/robotwin_approach_scene_change.png}{figures/stage_type_examples/libero_approach_scene_change.png}{Reassess the changed scene and approach the currently valid object or region.} &
\stageexamplecard{Contact--Wrong Pose}{figures/stage_type_examples/robotwin_contact_wrong_pose.png}{figures/stage_type_examples/libero_contact_wrong_pose.png}{Restore effective contact or grasp from an unsuitable robot--object pose.} &
\stageexamplecard{Move--Wrong Pose}{figures/stage_type_examples/robotwin_move_wrong_pose.png}{figures/stage_type_examples/libero_move_wrong_pose.png}{Correct the object or end-effector pose before continuing transport.} \\
\stageexamplecard{Place--Wrong Target}{figures/stage_type_examples/robotwin_place_wrong_target.png}{figures/stage_type_examples/libero_place_wrong_target.png}{Redirect the object from the wrong goal region to the instructed destination.} &
\stageexamplecard{Place--Wrong Pose}{figures/stage_type_examples/robotwin_place_wrong_pose.png}{figures/stage_type_examples/libero_place_wrong_pose.png}{Realign the object and robot before completing placement.} &
\stageexamplecard{Place--Scene Change}{figures/stage_type_examples/robotwin_place_scene_change.png}{figures/stage_type_examples/libero_place_scene_change.png}{Reassess the changed target configuration before completing placement.} \\
\bottomrule
\end{tabular}
}
\caption{\textbf{Representative recovery observations for the nine occupied
Stage--Type groups.} Each panel pairs RoboTwin (left) and LIBERO (right)
examples and summarizes the recovery required to continue the original task.}
\label{fig:app-stage-type-examples}
\end{figure*}

\subsection{Difficulty Screening and Replay Validation}
% 4 policies x 5 trials = 20；保留成功 2--18 次的场景
% 随机抽取 3 条 inference trajectory，seed 42，验证通过率 97.32%
% 说明困难状态可能放大 replay 误差，因此需要人工比对
Each candidate is evaluated five times by each of four screening policies, giving 20 recovery trials. We retain candidates with 2--18 successful trials, equivalent to an empirical recovery rate between 10\% and 90\%. The screening policies and source-specific construction procedures are listed in the main paper.

Difficult interaction states can be sensitive to small replay differences. Consequently, a prefix that executes successfully may still reconstruct a condition that differs from the one used during candidate screening. To detect such cases, we randomly select three of the screening inference trajectories for each retained candidate using seed 42. We replay the corresponding action prefixes and compare the reconstructed executions with the original trajectories around the recovery point.

The simulator-state and image criteria follow the replay-validation procedure defined in the main paper. Human review is used only to confirm that the reconstructed robot--scene relation and task progress remain behaviorally consistent. The resulting validation pass rate is 97.32\%. Candidates with inconsistent reconstruction, simulator failure, or invalid actions are removed before the train/test split is finalized.

\subsection{Corpus Composition}
% 两个平台各 800 train / 200 test
% train/test tasks、construction source、Natural source-policy 分布
% 不加入候选数据流数量
Each platform contains 1,000 scenarios with a fixed 800/200 train/test split. RoboTwin contains only Natural scenarios, whereas LIBERO combines Natural, Action-Perturbed, and Human-Constructed scenarios. Table~\ref{tab:app-corpus} summarizes the composition without including intermediate candidate-flow counts.

\begin{table}[b]
\centering
\small
\setlength{\tabcolsep}{2.5pt}
\renewcommand{\arraystretch}{1.02}
\begin{tabular*}{\columnwidth}
{@{\extracolsep{\fill}}llrrrrr@{}}
\toprule
Platform & Split & Scen. & Tasks & Natural & Pert. & Human \\
\midrule
RoboTwin & Train & 800 & 42 & 800 & --  & --  \\
         & Test  & 200 & 34 & 200 & --  & --  \\
LIBERO   & Train & 800 & 40 & 170 & 504 & 126 \\
         & Test  & 200 & 38 &  42 & 127 &  31 \\
\bottomrule
\end{tabular*}
\caption{\textbf{Corpus composition.} Pert. denotes Action-Perturbed scenarios;
dashes indicate construction sources not used on RoboTwin.}
\label{tab:app-corpus}
\end{table}

RoboTwin preserves the Natural source-policy distribution in both splits. The train/test counts are 169/24 for X-VLA, 112/37 for LingBot-VLA, 320/83 for SmolVLA, and 199/56 for $\pi_{0.5}$. The 42 LIBERO Natural test scenarios contain 16 $\pi_0$-source, 12 $\pi_{0.5}$-source, 9 Being-H0.5-source, and 5 UniFOLM-source scenarios. Fast-WAM construction uses Being-H0.5 rollouts and is included in the Being-H0.5 source group.

RoboTwin train and test sets share no sample identifier, JSON path, or exact instruction group. The 200 test scenarios form 183 trajectory clusters because 17 clusters contain two scenarios but different inference trajectories. LIBERO train and test sets share no Data-Path or complete episode key. 

The action-prefix replay protocol, policy initialization, and recovery horizon are defined in the main paper. In implementation, every candidate policy receives an independently reset and replayed environment instance. Its observation/action is initialized at the recovery point. Replay actions do not consume the environment steps. A trial terminates when the original task-success criteria is satisfied, the task-specific recovery environment steps reach the upper limit, or the simulator or policy adapter reports a runtime failure.

\section{Evaluation Implementation and Statistical Protocol}
% action-prefix replay、fresh history、完整 recovery horizon
% 各策略 repeat schedule、RSR、Macro-9、bootstrap
% runtime failure、Fast-WAM source 归入 Being-H0.5 source

Overall RSR follows the scenario-equal estimator defined in the main paper. Macro-9 RSR first averages scenario-level recovery rates within each of the nine Stage--Type groups and then gives the nine group rates equal weight. Macro-9 is therefore a category-balanced diagnostic, whereas Overall RSR preserves the empirical Stage--Type frequencies of the fixed test set.

RoboTwin confidence intervals use 10,000 trajectory-cluster bootstrap draws over its 183 test clusters. LIBERO intervals use 10,000 scenario-bootstrap draws over the 200 test scenarios. Paired policy contrasts jointly resample both policies' outcomes. Unless stated otherwise, pairwise intervals are not adjusted for multiple comparisons.

\section{Complete Recovery Results}
% 完整 initial-state / recovery 表
% paired contrasts 和置信区间
% 不再重复正文中的主要分析
This section reports the complete initial-state and recovery results. Table~\ref{tab:app-overall} gives the point estimates and 95\% confidence intervals for all 12 platform-policy evaluations on the fixed test splits. RoboTwin intervals use trajectory-cluster bootstrap, while LIBERO intervals use scenario bootstrap. The main paper provides the primary interpretation.

\begin{table}[t]
\centering
\small
\setlength{\tabcolsep}{1.5pt}
\begin{tabular*}{\columnwidth}{@{\extracolsep{\fill}}lcc@{}}
\toprule
Policy & Initial-state [95\% CI] & Recovery [95\% CI] \\
\midrule
\multicolumn{3}{@{}l}{\textit{RoboTwin}} \\
X-VLA       & 54.33 [47.60,60.86] & 49.83 [43.25,56.35] \\
LingBot-VLA & 64.33 [57.62,70.65] & 43.20 [36.60,49.57] \\
SmolVLA     & 42.67 [37.11,48.13] & 33.03 [27.66,38.65] \\
$\pi_{0.5}$ & 34.50 [29.81,39.27] & 32.60 [27.48,37.83] \\
Fast-WAM    & 81.00 [75.00,86.63] & 49.33 [42.12,56.46] \\
LingBot-VA  & \textbf{83.75} [78.71,88.32]
            & \textbf{59.50} [52.45,66.67] \\
\midrule
\multicolumn{3}{@{}l}{\textit{LIBERO}} \\
$\pi_0$       & 85.17 [81.17,89.00] & 37.80 [32.70,43.20] \\
$\pi_{0.5}$   & 94.17 [91.50,96.50]
              & \textbf{64.40} [59.20,69.50] \\
Being-H0.5    & 91.67 [88.33,94.67] & 46.80 [40.90,52.70] \\
UniFOLM       & \textbf{98.83} [97.83,99.67]
              & 48.00 [41.60,54.60] \\
Cosmos-Policy & 97.83 [96.50,99.00] & 52.10 [45.70,58.40] \\
Fast-WAM      & 96.83 [94.50,98.67] & 54.00 [47.80,60.10] \\
\bottomrule
\end{tabular*}
\caption{\textbf{Complete initial-state and recovery results on the fixed test
splits (\%).} Bold marks the highest point estimate in each platform and
evaluation condition.}
\label{tab:app-overall}
\end{table}

Table~\ref{tab:app-macro9} reports Macro-9 RSR, which weights the nine
Stage--Type groups equally; Table~\ref{tab:app-overall} reports
scenario-weighted Overall RSR.

\begin{table}[t]
\centering
\small
\setlength{\tabcolsep}{2.5pt}
\begin{tabular*}{\columnwidth}{@{\extracolsep{\fill}}lrlr@{}}
\toprule
RoboTwin policy & Macro-9 & LIBERO policy & Macro-9 \\
\midrule
X-VLA       & \RTXMacro        & $\pi_0$       & \LBPiMacro \\
LingBot-VLA & \RTLingVLAMacro  & $\pi_{0.5}$   & \textbf{\LBPiFiveMacro} \\
SmolVLA     & \RTSmolMacro     & Being-H0.5    & \LBBeingMacro \\
$\pi_{0.5}$ & \RTPiFiveMacro   & UniFOLM       & \LBUniMacro \\
Fast-WAM    & \textbf{\RTFastMacro} & Cosmos-Policy & \LBCosmosMacro \\
LingBot-VA  & \RTLingVAMacro   & Fast-WAM      & \LBFastMacro \\
\bottomrule
\end{tabular*}
\caption{\textbf{Macro-9 RSR on the fixed test splits (\%), computed from unrounded
scenario-level rates.} Best values are bold.}
\label{tab:app-macro9}
\end{table}

\subsection{Paired Policy Contrasts} 

Table~\ref{tab:app-paired-contrasts} reports all paired policy contrasts on the same test units. Each entry gives policy \(A\) minus policy \(B\), so a positive value favors \(A\). The intervals are not adjusted for multiple comparisons.

For RoboTwin, X, L, S, P, F, and V denote X-VLA, LingBot-VLA, SmolVLA, $\pi_{0.5}$, Fast-WAM, and LingBot-VA. For LIBERO, P0, P5, B, U, C, and F denote $\pi_0$, $\pi_{0.5}$, Being-H0.5, UniFOLM, Cosmos-Policy, and Fast-WAM.

\begin{table}[t]
\centering
\small
\setlength{\tabcolsep}{1.5pt}
\begin{tabular*}{\columnwidth}{@{\extracolsep{\fill}}lrr@{}}
\toprule
Pair \(A-B\) &
\shortstack{Initial-state \(\Delta\)\\(95\% CI)} &
\shortstack{Recovery \(\Delta\)\\(95\% CI)} \\
\midrule
\multicolumn{3}{@{}l}{\textit{RoboTwin}} \\
\(X-L\) & -10.00 [-19.24,-0.51] & 6.63 [-3.45,16.73] \\
\(X-S\) & 11.67 [3.85,19.50] & 16.80 [8.38,25.08] \\
\(X-P\) & 19.83 [11.68,27.81] & 17.23 [8.16,26.21] \\
\(X-F\) & -26.67 [-34.97,-18.26] & 0.50 [-9.02,10.23] \\
\(X-V\) & -29.42 [-37.35,-21.49] & -9.67 [-18.77,-0.84] \\
\(L-S\) & 21.67 [12.41,30.36] & 10.17 [1.51,18.41] \\
\(L-P\) & 29.83 [21.62,37.56] & 10.60 [1.78,19.32] \\
\(L-F\) & -16.67 [-23.42,-10.03] & -6.13 [-14.35,2.26] \\
\(L-V\) & -19.42 [-25.50,-13.50] & -16.30 [-26.33,-6.57] \\
\(S-P\) & 8.17 [1.33,14.93] & 0.43 [-5.95,7.04] \\
\(S-F\) & -38.33 [-45.65,-30.69] & -16.30 [-24.38,-8.23] \\
\(S-V\) & -41.08 [-47.97,-34.02] & -26.47 [-35.74,-17.35] \\
\(P-F\) & -46.50 [-53.33,-39.26] & -16.73 [-24.35,-9.10] \\
\(P-V\) & -49.25 [-55.53,-42.64] & -26.90 [-36.50,-17.66] \\
\(F-V\) & -2.75 [-7.71,2.26] & -10.17 [-20.41,0.00] \\
\midrule
\multicolumn{3}{@{}l}{\textit{LIBERO}} \\
\(P0-P5\) & -9.00 [-12.67,-5.67] & -26.60 [-33.40,-19.90] \\
\(P0-B\)  & -6.50 [-11.33,-1.83] & -9.00 [-16.30,-1.60] \\
\(P0-U\)  & -13.67 [-17.50,-10.00] & -10.20 [-17.80,-2.60] \\
\(P0-C\)  & -12.67 [-16.83,-8.67] & -14.30 [-22.00,-6.60] \\
\(P0-F\)  & -11.67 [-15.67,-8.00] & -16.20 [-23.50,-8.90] \\
\(P5-B\)  & 2.50 [-1.50,6.50] & 17.60 [10.00,25.10] \\
\(P5-U\)  & -4.67 [-7.50,-2.33] & 16.40 [9.10,23.80] \\
\(P5-C\)  & -3.67 [-6.67,-1.00] & 12.30 [4.30,20.20] \\
\(P5-F\)  & -2.67 [-5.83,0.50] & 10.40 [2.40,18.30] \\
\(B-U\)   & -7.17 [-10.50,-4.00] & -1.20 [-9.00,6.70] \\
\(B-C\)   & -6.17 [-9.83,-2.83] & -5.30 [-13.20,2.70] \\
\(B-F\)   & -5.17 [-9.00,-1.33] & -7.20 [-14.30,-0.20] \\
\(U-C\)   & 1.00 [-0.33,2.50] & -4.10 [-11.40,3.20] \\
\(U-F\)   & 2.00 [-0.17,4.50] & -6.00 [-12.60,0.90] \\
\(C-F\)   & 1.00 [-1.33,3.67] & -1.90 [-8.30,4.70] \\
\bottomrule
\end{tabular*}
\caption{\textbf{All paired policy contrasts in percentage points.} RoboTwin
intervals use trajectory-cluster bootstrap; LIBERO intervals use paired
scenario bootstrap.}
\label{tab:app-paired-contrasts}
\end{table}

\section{Comparison with LIBERO-Plus Preset-OOD Evaluation}

\vpara{LIBERO-Plus.}
LIBERO-Plus extends the original LIBERO benchmark with controlled perturbations in seven dimensions: camera viewpoint, robot initial state, language instruction, lighting, background texture, sensor noise, and object layout~\citep{fei2025liberoplus}. These perturbations are specified before a rollout begins. The policy then controls the complete episode from the perturbed initialization. LIBERO-Plus therefore measures robustness to predefined visual, linguistic, geometric, and environmental distribution shifts.

\vpara{Difference from RoboRecover.}
RoboRecover evaluates a different condition. Its recovery point is reached after previous robot actions and contacts have changed the scene and produced partial task progress. The candidate policy must interpret this intermediate condition, correct the deviation, and continue the original task.

Thus, LIBERO-Plus asks whether a policy can complete a task when an external variation is introduced before execution. RoboRecover asks whether a policy can continue after execution itself has produced an off-nominal state. The former emphasizes robustness to preset OOD conditions, while the latter emphasizes reassessment and correction during task execution.

\vpara{Overlapping policies.}
RoboRecover contains 12 platform-policy entries but 10 unique model families, because $\pi_{0.5}$ and Fast-WAM are evaluated on both platforms. Published LIBERO-Plus results are available for five of the six policies in our LIBERO evaluation. Table~\ref{tab:app-libero-plus-overlap} compares these results with RoboRecover. LIBERO-Plus values are taken from published evaluations rather than rerun in our environment.

\begin{table}[t]
\centering
\small
\setlength{\tabcolsep}{2.5pt}
\begin{tabular*}{\columnwidth}{@{\extracolsep{\fill}}lrrr@{}}
\toprule
Model &
\shortstack{LIBERO-Plus\\Total} &
\shortstack{RoboRecover\\Initial-state} &
\shortstack{RoboRecover\\RSR} \\
\midrule
$\pi_0$       & 53.6 (69.4 rerun) & \LBPiSeed    & \LBPiRSR \\
$\pi_{0.5}$   & \textbf{85.7}              & \LBPiFiveSeed & \textbf{\LBPiFiveRSR} \\
Being-H0.5    & 78.5$^\dagger$    & \LBBeingSeed & \LBBeingRSR \\
UniFOLM       & --                 & \textbf{\LBUniSeed}   & \LBUniRSR \\
Cosmos-Policy & \underline{82.2}              & \underline{\LBCosmosSeed} & \LBCosmosRSR \\
Fast-WAM      & 51.5              & \LBFastSeed  & \underline{\LBFastRSR} \\
\bottomrule
\end{tabular*}

\par\smallskip

\begin{tabular*}{\columnwidth}{@{\extracolsep{\fill}}llrp{0.40\columnwidth}@{}}
\toprule
Model & RR platform & L-Plus & Comparability \\
\midrule
X-VLA &
RoboTwin &
71.4 &
LIBERO checkpoint; model-family comparison only. \\

SmolVLA &
RoboTwin &
45.37$^\ddagger$ &
Separately trained LIBERO checkpoint; model-family comparison only. \\

LingBot-VLA &
RoboTwin &
-- &
No published aggregate result identified. \\

LingBot-VA &
RoboTwin &
-- &
RoboTwin 2.0-Plus only; no LIBERO-Plus result. \\
\bottomrule
\end{tabular*}
\caption{\textbf{Published LIBERO-Plus results and RoboRecover results for
overlapping model families (\%).} The $\pi_0$ value in parentheses is the
JAX rerun reported by \citet{zhang2026wamrobustness}. $^\dagger$Being-H0.5
is reported by \citet{luo2026beingh07}. $^\ddagger$SmolVLA is reported
under a separate LIBERO training and evaluation setup by
\citet{li2026corridorvla}. Dashes indicate that no directly reported
aggregate LIBERO-Plus result was identified.}
\label{tab:app-libero-plus-overlap}
\end{table}

\vpara{The two evaluations produce different policy orderings.}
Among the four policies evaluated together in
\citet{zhang2026wamrobustness}, the LIBERO-Plus order is
\[
\pi_{0.5} > \text{Cosmos-Policy} > \pi_0 > \text{Fast-WAM}.
\]
Their RoboRecover order is
\[
\pi_{0.5} > \text{Fast-WAM} > \text{Cosmos-Policy} > \pi_0.
\]
The ordering is unchanged if the reported $\pi_0$ JAX rerun is used instead of its original LIBERO-Plus value.

Fast-WAM shows the clearest reversal. It has the lowest LIBERO-Plus result among these four policies at 51.5\%, but the second-highest RoboRecover rate at \LBFastRSR\%. Cosmos-Policy shows the opposite tendency: it reaches 82.2\% on LIBERO-Plus but \LBCosmosRSR\% on RoboRecover. These results suggest that robustness to predefined perturbations does not directly determine the ability to interpret and repair an execution-induced intermediate state.

The result is not that one benchmark is harder or more complete than the other. They expose different failure conditions. A policy may tolerate changes in camera, lighting, layout, or robot initialization but still fail after its own interaction has altered object relations and task progress. Conversely, a policy may recover effectively from an intermediate deviation while remaining sensitive to externally imposed visual or geometric shifts.

Ranking differences alone do not prove which model component causes these behaviors, because checkpoints, training data, and evaluation implementations also differ. They do provide evidence that LIBERO-Plus and RoboRecover scores are not interchangeable. Preset-OOD evaluation measures robustness to conditions defined before execution, whereas recovery evaluation measures the ability to reassess and correct the consequences of prior interaction.

\section{Stage--Type and Source-Conditioned Results}
% 九个 Stage--Type 完整结果
% source-policy x recovery-policy 矩阵
% construction-source 和 task-suite 结果
% oracle 仅说明不同策略成功场景不同，不称为可部署方法
This section reports the complete results conditioned on Recovery Stage, Deviation Type, construction source, task suite, and Natural source policy. These tables describe observed test-set behavior. They are not claims about the internal causes of policy behavior.

Tables~\ref{tab:app-rt-stage} and~\ref{tab:app-lib-stage} report all nine nonempty Stage--Type groups. \(N\) is the number of scenarios in each group. Bold marks the largest point estimate in a row. Results from very small groups should be treated as descriptive.

\begin{table}[t]
\centering
\small
\setlength{\tabcolsep}{2.2pt}
\begin{tabular*}{\columnwidth}{@{\extracolsep{\fill}}lrrr@{}}
\toprule
RoboTwin Stage--Type (\(N\)) & X-VLA & LingBot-VLA & SmolVLA \\
\midrule
Approach--Scene Change (4)    & 8.33 & 33.33 & 0.00 \\
Approach--Stage Rollback (24) & 46.67 & 26.39 & 16.11 \\
Approach--Wrong Pose (44)     & 66.36 & 41.06 & 30.91 \\
Approach--Wrong Target (29)   & 49.43 & 38.62 & 16.09 \\
Contact--Wrong Pose (21)      & 43.81 & \textbf{65.71} & 35.87 \\
Move--Wrong Pose (35)         & 53.52 & 33.33 & 46.10 \\
Place--Scene Change (2)       & 16.67 & 16.67 & 66.67 \\
Place--Wrong Pose (23)        & 34.78 & 53.62 & 54.78 \\
Place--Wrong Target (18)      & 46.30 & \textbf{62.96} & 35.19 \\
\bottomrule
\end{tabular*}

\par\smallskip

\begin{tabular*}{\columnwidth}{@{\extracolsep{\fill}}lrrr@{}}
\toprule
RoboTwin Stage--Type (\(N\)) & \(\pi_{0.5}\) & Fast-WAM & LingBot-VA \\
\midrule
Approach--Scene Change (4)    & 18.33 & \textbf{75.00} & 50.00 \\
Approach--Stage Rollback (24) & 19.72 & 29.17 & \textbf{62.50} \\
Approach--Wrong Pose (44)     & 22.58 & 50.00 & \textbf{86.36} \\
Approach--Wrong Target (29)   & 22.76 & 41.38 & \textbf{62.07} \\
Contact--Wrong Pose (21)      & 31.75 & 52.38 & 52.38 \\
Move--Wrong Pose (35)         & 58.48 & 48.57 & \textbf{71.43} \\
Place--Scene Change (2)       & 66.67 & \textbf{83.33} & 50.00 \\
Place--Wrong Pose (23)        & 47.54 & \textbf{65.22} & 21.74 \\
Place--Wrong Target (18)      & 21.11 & 55.56 & 22.22 \\
\bottomrule
\end{tabular*}
\caption{\textbf{Complete RoboTwin recovery rates across Stage--Type groups (\%).}
The two Scene Change groups contain only four and two scenarios.}
\label{tab:app-rt-stage}
\end{table}

\begin{table}[t]
\centering
\small
\setlength{\tabcolsep}{2.2pt}
\begin{tabular*}{\columnwidth}{@{\extracolsep{\fill}}lrrr@{}}
\toprule
LIBERO Stage--Type (\(N\)) & \(\pi_0\) & \(\pi_{0.5}\) & Being-H0.5 \\
\midrule
Approach--Scene Change (28)   & 12.1 & \textbf{50.0} & 24.3 \\
Approach--Stage Rollback (21) & 39.0 & \textbf{77.1} & 35.2 \\
Approach--Wrong Pose (49)     & 42.4 & 59.6 & 42.4 \\
Approach--Wrong Target (25)   & 40.8 & 70.4 & 58.4 \\
Contact--Wrong Pose (12)      & 30.0 & \textbf{55.0} & 48.3 \\
Move--Wrong Pose (16)         & 53.8 & 71.2 & 63.7 \\
Place--Scene Change (11)      & 43.6 & \textbf{81.8} & 34.5 \\
Place--Wrong Pose (31)        & 46.5 & 61.3 & \textbf{62.6} \\
Place--Wrong Target (7)       & 22.9 & \textbf{82.9} & 68.6 \\
\bottomrule
\end{tabular*}

\par\smallskip

\begin{tabular*}{\columnwidth}{@{\extracolsep{\fill}}lrrr@{}}
\toprule
LIBERO Stage--Type (\(N\)) & UniFOLM & Cosmos & Fast-WAM \\
\midrule
Approach--Scene Change (28)   & 27.9 & 37.1 & 36.4 \\
Approach--Stage Rollback (21) & 29.5 & 49.5 & 43.8 \\
Approach--Wrong Pose (49)     & 46.9 & \textbf{64.1} & 58.4 \\
Approach--Wrong Target (25)   & 55.2 & 53.6 & \textbf{72.8} \\
Contact--Wrong Pose (12)      & 41.7 & 36.7 & 43.3 \\
Move--Wrong Pose (16)         & 56.2 & 55.0 & \textbf{72.5} \\
Place--Scene Change (11)      & 67.3 & 50.9 & 21.8 \\
Place--Wrong Pose (31)        & 61.9 & 51.0 & 54.2 \\
Place--Wrong Target (7)       & 65.7 & 57.1 & \textbf{82.9} \\
\bottomrule
\end{tabular*}
\caption{\textbf{Complete LIBERO recovery rates across Stage--Type groups (\%).}
Cosmos denotes Cosmos-Policy. The Place--Wrong Target group contains seven
scenarios and is reported descriptively.}
\label{tab:app-lib-stage}
\end{table}

\subsection{Construction-Source and Task-Suite Results}
RoboTwin scenarios are constructed from Natural rollouts. LIBERO additionally contains Action-Perturbed and Human-Constructed scenarios. Table~\ref{tab:app-construction-suite} reports LIBERO recovery rates by construction source and task suite. These results describe performance on the fixed test composition rather than the prevalence or causal difficulty of each condition.

\begin{table}[t]
\centering
\small
\setlength{\tabcolsep}{2.2pt}
\begin{tabular*}{\columnwidth}{@{\extracolsep{\fill}}lrrr@{}}
\toprule
Group (\(N\)) & \(\pi_0\) & \(\pi_{0.5}\) & Being-H0.5 \\
\midrule
\multicolumn{4}{@{}l}{\textit{Construction source}} \\
Human-constructed (31) & 44.5 & 70.3 & 38.7 \\
Natural (42)            & 48.6 & 58.1 & 55.7 \\
Action-perturbed (127)  & 32.6 & 65.0 & 45.8 \\
\addlinespace[2pt]
\multicolumn{4}{@{}l}{\textit{Task suite}} \\
LIBERO-Spatial (28) & 40.7 & 74.3 & 39.3 \\
LIBERO-Object (57)  & 33.0 & 60.7 & 49.5 \\
LIBERO-Goal (57)    & 35.4 & 67.0 & 42.8 \\
LIBERO-10 (58)      & 43.4 & 60.7 & 51.7 \\
\bottomrule
\end{tabular*}

\par\smallskip

\begin{tabular*}{\columnwidth}{@{\extracolsep{\fill}}lrrr@{}}
\toprule
Group (\(N\)) & UniFOLM & Cosmos & Fast-WAM \\
\midrule
\multicolumn{4}{@{}l}{\textit{Construction source}} \\
Human-constructed (31) & 50.3 & 43.9 & 41.3 \\
Natural (42)            & 59.5 & 70.5 & 67.1 \\
Action-perturbed (127)  & 43.6 & 48.0 & 52.8 \\
\addlinespace[2pt]
\multicolumn{4}{@{}l}{\textit{Task suite}} \\
LIBERO-Spatial (28) & 51.4 & 60.7 & 47.9 \\
LIBERO-Object (57)  & 46.7 & 57.2 & 55.8 \\
LIBERO-Goal (57)    & 48.4 & 43.5 & 56.5 \\
LIBERO-10 (58)      & 47.2 & 51.4 & 52.8 \\
\bottomrule
\end{tabular*}
\caption{\textbf{LIBERO recovery rates by construction source and task suite
(\%).}}
\label{tab:app-construction-suite}
\end{table}

The construction-source results vary by recovery policy. For example, $\pi_{0.5}$ is strongest on Human-Constructed and Action-Perturbed scenarios, whereas Cosmos-Policy is strongest on Natural scenarios. The task-suite results likewise show that no suite has the same ordering for every policy. These comparisons should be read with the group sizes reported in the table.

\subsection{Natural Source-Policy Results}
Natural scenarios retain the policy whose rollout produced the deviation. Table~\ref{tab:app-source-matrices} fixes this source policy by row and gives the evaluated recovery policy by column. Italics mark reevaluation of the source policy, and bold marks the largest observed recovery rate in each row.

\begin{table}[t]
\centering
\small
\setlength{\tabcolsep}{2.2pt}
\begin{tabular*}{\columnwidth}{@{\extracolsep{\fill}}lrrr@{}}
\toprule
RoboTwin source (\(N\)) & X-VLA & LingBot-VLA & SmolVLA \\
\midrule
X-VLA (24)         & \textit{17.78} & 42.22 & \textbf{58.06} \\
LingBot-VLA (37)   & 59.10 & \textit{19.82} & 48.83 \\
SmolVLA (83)       & 43.21 & 57.67 & \textit{17.27} \\
\(\pi_{0.5}\) (56) & \textbf{67.26} & 37.62 & 35.24 \\
\bottomrule
\end{tabular*}

\par\smallskip

\begin{tabular*}{\columnwidth}{@{\extracolsep{\fill}}lrrr@{}}
\toprule
RoboTwin source (\(N\)) & \(\pi_{0.5}\) & Fast-WAM & LingBot-VA \\
\midrule
X-VLA (24)         & 47.78 & 40.28 & 54.17 \\
LingBot-VLA (37)   & 50.09 & 54.05 & \textbf{70.27} \\
SmolVLA (83)       & 32.13 & \textbf{61.45} & 60.24 \\
\(\pi_{0.5}\) (56) & \textit{15.24} & 32.14 & 53.57 \\
\bottomrule
\end{tabular*}

\par\smallskip

\begin{tabular*}{\columnwidth}{@{\extracolsep{\fill}}lrrr@{}}
\toprule
LIBERO source (\(N\)) & \(\pi_0\) & \(\pi_{0.5}\) & Being-H0.5 \\
\midrule
\(\pi_0\) (16)     & \textit{31.25} & 60.00 & 63.75 \\
\(\pi_{0.5}\) (12) & 53.33 & \textit{30.00} & 50.00 \\
Being-H0.5 (9)     & 73.33 & \textbf{80.00} & \textit{64.44} \\
UniFOLM (5)        & 48.00 & \textbf{80.00} & 28.00 \\
\bottomrule
\end{tabular*}

\par\smallskip

\begin{tabular*}{\columnwidth}{@{\extracolsep{\fill}}lrrr@{}}
\toprule
LIBERO source (\(N\)) & UniFOLM & Cosmos & Fast-WAM \\
\midrule
\(\pi_0\) (16)     & 62.50 & \textbf{70.00} & 63.75 \\
\(\pi_{0.5}\) (12) & 63.33 & \textbf{91.67} & 90.00 \\
Being-H0.5 (9)     & 42.22 & 46.67 & 44.44 \\
UniFOLM (5)        & \textit{72.00} & 64.00 & 64.00 \\
\bottomrule
\end{tabular*}
\caption{\textbf{Complete Natural source-policy by recovery-policy matrices
(\%). }Cosmos denotes Cosmos-Policy. Rows with small \(N\) are descriptive.}
\label{tab:app-source-matrices}
\end{table}

The source matrices show that the policy producing a deviation is not necessarily the policy with the highest recovery rate from that deviation. This comparison is possible because all recovery policies are evaluated on the same reconstructed scenarios within each source group.

For each test scenario, we also compute the largest empirical recovery rate observed among the six evaluated policies. This retrospective value reaches 95.63\% on RoboTwin and 91.80\% on LIBERO, compared with 59.50\% and 64.40\% for the best fixed policy on each platform. The corresponding differences are 36.13 and 27.40 points. No RoboTwin scenario and one LIBERO scenario have zero observed success across all evaluated policies. 

This calculation uses test outcomes to select a policy separately for every scenario. It is therefore not a deployable policy-selection method. It only shows that the evaluated policies succeed on different sets of recovery scenarios.

\section{Recovery Mechanism Studies}

The primary benchmark evaluates a fixed policy without additional assistance. This section studies how the training split can support recovery mechanisms. These experiments are exploratory and are not part of the primary direct-recovery leaderboard.

\subsection{Monitor Training and Triggering}
% RoboTwin：799 IDs，719/80 split，320k train，8k balanced val，16k natural val
% 报告完整 Accuracy/Precision/Recall/F1/Macro-F1/FPR 与混淆矩阵
% X-VLA/LingBot-VLA trigger rate = 76%/83%
% 单帧 1 触发、每个 episode 最多一次、replay 阶段不监测
We construct the RoboTwin monitor data from 800 unique scenario IDs.  We assign 720 IDs to training and 80 IDs to validation.  All frames associated with the same ID remain in the same split.  This prevents frames from the same scenario from appearing in both training and validation.

Frames before the recovery point, excluding the recovery point itself, are labeled 0.  The recovery-point frame and subsequent frames are labeled 1.  The training set is balanced by sampling the same number of frames from both classes.  We use two views of the held-out IDs: a balanced validation set and a natural validation set that preserves the observed label distribution.

Table~\ref{tab:app-monitor-data} gives the scenario and frame counts for the training split and both validation views.

\begin{table}[t]
\centering
\small
\setlength{\tabcolsep}{3pt}
\begin{tabular*}{\columnwidth}{@{\extracolsep{\fill}}lrrrr@{}}
\toprule
Split & IDs & Frames & Label 0 & Label 1 \\
\midrule
Training       & 720 & 320,000 & 160,000 & 160,000 \\
Balanced val.  & 80  & 8,000   & 4,000   & 4,000 \\
Natural val.   & 80  & 16,000  & 5,831   & 10,169 \\
\bottomrule
\end{tabular*}
\caption{\textbf{RoboTwin monitor training and validation data.} The two validation
sets use different frame-sampling distributions from the same 80 held-out
scenario IDs.}
\label{tab:app-monitor-data}
\end{table}

The monitor is trained for two epochs and 5,000 optimizer steps. We select the merged checkpoint at step 5,000. Table~\ref{tab:app-monitor-metrics} reports the complete validation results. Precision, recall, and F1 refer to label 1.

\begin{table}[t]
\centering
\small
\setlength{\tabcolsep}{2pt}
\begin{tabular*}{\columnwidth}{@{\extracolsep{\fill}}lrrrrrr@{}}
\toprule
Validation & Acc. & Prec. & Recall & F1 & Macro-F1 & FPR \\
\midrule
Balanced & 85.49 & 86.11 & 84.63 & 85.36 & 85.49 & 13.65 \\
Natural  & 86.42 & 91.99 & 86.14 & 88.97 & 85.66 & 13.09 \\
\bottomrule
\end{tabular*}

\par\smallskip

\begin{tabular*}{\columnwidth}{@{\extracolsep{\fill}}lrrrr@{}}
\toprule
Validation & TN & FP & FN & TP \\
\midrule
Balanced & 3,454 & 546 & 615   & 3,385 \\
Natural  & 5,068 & 763 & 1,409 & 8,760 \\
\bottomrule
\end{tabular*}
\caption{\textbf{RoboTwin monitor validation results (\%).} The lower panel gives
the corresponding confusion matrices. FPR is the false-positive rate.}
\label{tab:app-monitor-metrics}
\end{table}

Replay actions are not monitored. Monitoring begins when policy inference starts from the recovery point and is performed at every environment step. The output is parsed strictly as 0 or 1. A single prediction of 1 activates the intervention. Each episode can trigger at most once, and monitoring stops after the trigger. 

When monitoring X-VLA, 152 of 200 test episodes trigger, giving an episode-level trigger rate of 76.0\%. For LingBot-VLA, 166 of 200 episodes trigger, giving a rate of 83.0\%. These rates measure whether an intervention is activated at least once.

\subsection{Reversion Results}
% horizon <500 回退30步，否则60步
% X-VLA fixed Reversion 54.5%，monitor-triggered 59.0%
% 仅作探索性结果
Reversion returns execution to an earlier point with less accumulated deviation. The return distance depends on the task horizon. We return 30 environment steps when the maximum horizon is below 500 steps and 60 steps otherwise. When fewer preceding steps are available, the actual distance is limited to the available execution history.

In the fixed setting, Reversion is applied directly at the recovery point. In the monitor-triggered setting, X-VLA first continues from the recovery point. The first positive monitor output then activates Reversion, after which X-VLA resumes control.

On the RoboTwin test split, fixed Reversion with X-VLA reaches 54.5\% recovery. Monitor-triggered Reversion reaches 59.0\%. These results use one base policy and one Reversion configuration. They therefore provide an exploratory demonstration rather than a general comparison of recovery mechanisms.

\subsection{LIBERO Monitor and Human Intervention}
% Qwen3-VL-4B，262,669 frames，LoRA rank16，lr 5e-5
% ckpt-4104；不报告不存在的 precision/recall/F1
% HIL N=50：direct takeover 82%，monitor-triggered 88%
The LIBERO monitor fine-tunes Qwen3-VL-4B-Instruct on 262,669 labeled frames. The data contain 102,357 label-0 frames and 160,312 label-1 frames. Training uses LoRA with rank 16, a batch size of 4, gradient accumulation over 8 batches, and an effective batch size of 32. The learning rate is \(5\times10^{-5}\).

We additionally test human intervention on 50 LIBERO scenarios. In the direct setting, the human operator takes control from the recovery point and succeeds on 82\% of the scenarios. In the monitor-triggered setting, the base policy acts until the first positive monitor output, after which the human takes control. This setting succeeds on 88\% of the scenarios. Because this study contains only 50 scenarios and uses human control, the results are reported as an exploratory reference rather than a policy benchmark.

\subsection{Correction Data and Training}
% 两次 run：522->78->step2000；156->29->step2999
% donor 最终分布：15/17/46；6/8/15
% 当前配置：action horizon10、batch256、AdamW、EMA .999、
% warmup200、lr 5e-5、3000 steps
% 不将两个 run 相加；不报告尚未对应 checkpoint 的结果
The Correction controller is trained from successful trajectories produced by alternative donor policies. Each donor trajectory is first evaluated in four trials. We retain trajectories that succeed in at least three trials and complete all scheduled trials. The retained list is then deduplicated by its source record and checked for valid training trajectories.

The current training initializes from the official LIBERO $\pi_{0.5}$ checkpoint and uses an action horizon of 10. Training uses a batch size of 256 and AdamW with gradient-norm clipping at 1.0. The learning rate is \(5\times10^{-5}\), with 200 warmup steps and a 3,000-step schedule. Exponential moving averaging uses a coefficient of 0.999. 

\end{document}